\documentclass[10pt,twocolumn,letterpaper]{article}

\usepackage{iccv}
\usepackage{times}
\usepackage{epsfig}
\usepackage{graphicx}
\usepackage{amsmath}
\usepackage{amssymb}

\usepackage{booktabs}
\usepackage{siunitx}
\usepackage{enumitem}
\usepackage{caption} 
\usepackage[pagebackref=true,breaklinks=true,letterpaper=true,colorlinks,bookmarks=false]{hyperref}

\usepackage[capitalize]{cleveref}
\crefname{section}{Sec.}{Secs.}
\Crefname{section}{Section}{Sections}
\Crefname{table}{Table}{Tables}
\crefname{table}{Tab.}{Tabs.}

\iccvfinalcopy 

\ificcvfinal\fi

\begin{document}

\title{LidarCLIP or: How I Learned to Talk to Point Clouds}


\author{Georg Hess$^{\dagger,1,2}$
\quad
Adam Tonderski$^{\dagger,1,3}$
\\
Christoffer Petersson$^{1,2}$
\quad
Kalle Åström$^{3}$
\quad
Lennart Svensson$^{2}$
\\
\normalsize$^1$Zenseact \hspace{1cm} $^2$Chalmers University of Technology \hspace{1cm} $^3$Lund University\\
{\tt\small \{first.last\}@zenseact.com} \hspace{0.5cm} \tt\small lennart.svensson@chalmers.se \hspace{0.5cm}   \tt\small karl.astrom@math.lth.se 
}
\maketitle

\begin{abstract}
    Research connecting text and images has recently seen several breakthroughs, with models like CLIP, DALL·E 2, and Stable Diffusion. However, the connection between text and other visual modalities, such as lidar data, has received less attention, prohibited by the lack of text-lidar datasets. In this work, we propose LidarCLIP, a mapping from automotive point clouds to a pre-existing CLIP embedding space. Using image-lidar pairs, we supervise a point cloud encoder with the image CLIP embeddings, effectively relating text and lidar data with the image domain as an intermediary. We show the effectiveness of LidarCLIP by demonstrating that lidar-based retrieval is generally on par with image-based retrieval, but with complementary strengths and weaknesses. By combining image and lidar features, we improve upon both single-modality methods and enable a targeted search for challenging detection scenarios under adverse sensor conditions. We also explore zero-shot classification and show that LidarCLIP outperforms existing attempts to use CLIP for point clouds by a large margin. Finally, we leverage our compatibility with CLIP to explore a range of applications, such as point cloud captioning and lidar-to-image generation, without any additional training. Code and pre-trained models are available at 
    \href{https://github.com/atonderski/lidarclip}{github.com/atonderski/lidarclip}.
\end{abstract}

\def\thefootnote{$\dagger$}\footnotetext{These authors contributed equally to this work.}\def\thefootnote{\arabic{footnote}}

\section{Introduction}
\label{sec:intro}
Connecting natural language processing (NLP) and computer vision (CV) has been a long-standing challenge in the research community. Recently, OpenAI released CLIP \cite{CLIP}, a model trained on 400 million web-scraped text-image pairs, that produces powerful text and image representations. Besides impressive zero-shot classification performance, CLIP enables interaction with the image domain in a diverse and intuitive way by using human language. These capabilities have resulted in a surge of work building upon the CLIP embeddings within multiple applications, such as image captioning \cite{clipcap}, image retrieval \cite{baldrati2022effective,hendriksen2022extending}, semantic segmentation \cite{zhou2022extract}, text-to-image generation \cite{dalle2,rombach2022high}, and referring image segmentation \cite{luddecke2022image,wang2022cris}.

\begin{figure}[tpb]
    \centering
    \includegraphics[width=\linewidth]{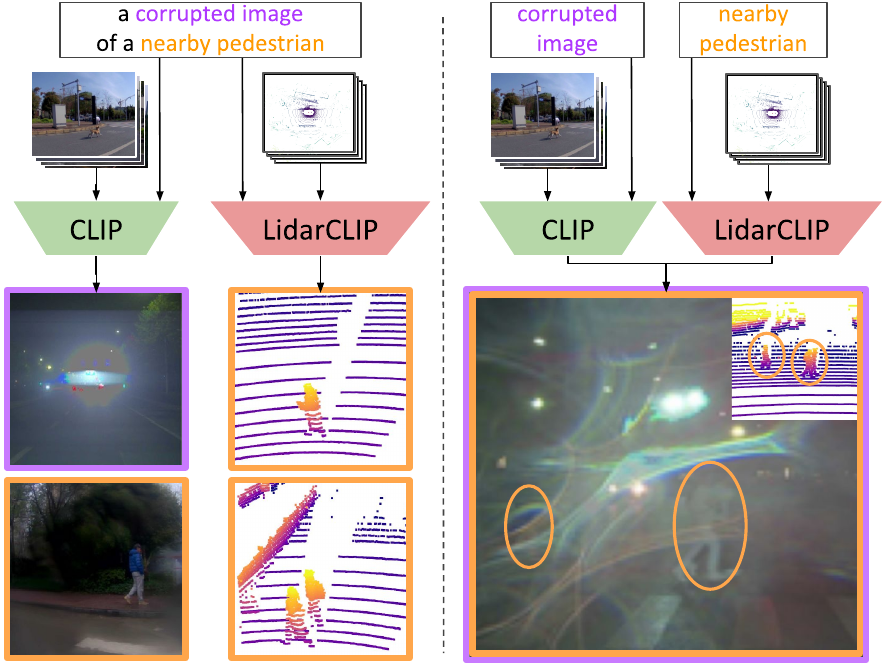}
    \caption{Like CLIP, LidarCLIP has many applications, including retrieval for data curation. Here, we demonstrate that the two can be combined through different queries to retrieve potentially safety-critical scenes that a camera-based system may handle poorly. Such scenes are nearly impossible to retrieve with a single modality.
    }
    \label{fig:first_page}
\end{figure}

While most works trying to bridge the gap between NLP and CV have focused on a single visual modality, namely images, other visual modalities, such as lidar point clouds, have received far less attention. Existing attempts to connect NLP and point clouds are often limited to a single application \cite{chen2020scanrefer,rozenberszki2022language,zhao20213dvg} or designed for synthetic data \cite{zhang2022pointclip}. This is a natural consequence due to the lack of large-scale text-lidar datasets required for training flexible models such as CLIP in a new domain. However, it has been shown that the CLIP embedding space can be extended to new languages \cite{carlssoncross} and new modalities, such as audio \cite{wu2022wav2clip}, without the need for huge datasets and extensive computational resources. This raises the question if such techniques can be applied to point clouds as well, and consequently open up a body of research on point cloud understanding, similar to what has emerged for images \cite{wang2022cris,zhou2022extract}.

We propose LidarCLIP, a method to connect the CLIP embedding space to the lidar point cloud domain. While combined text and point cloud datasets are not easily accessible, many robotics applications capture images and point clouds simultaneously. One example is autonomous driving, where data is both openly available and large scale. To this end, we supervise a lidar encoder with a frozen CLIP image encoder using pairs of images and point clouds from the large-scale automotive dataset ONCE \cite{mao2021one}. This way, the image encoder's rich and diverse semantic understanding is transferred to the point cloud domain. At inference, we can compare LidarCLIP's embedding of a point cloud with the embeddings from either CLIP's text encoder, image encoder, or both, enabling various applications.


\begin{figure*}[thb]
    \centering
    \includegraphics[width=0.9\linewidth]{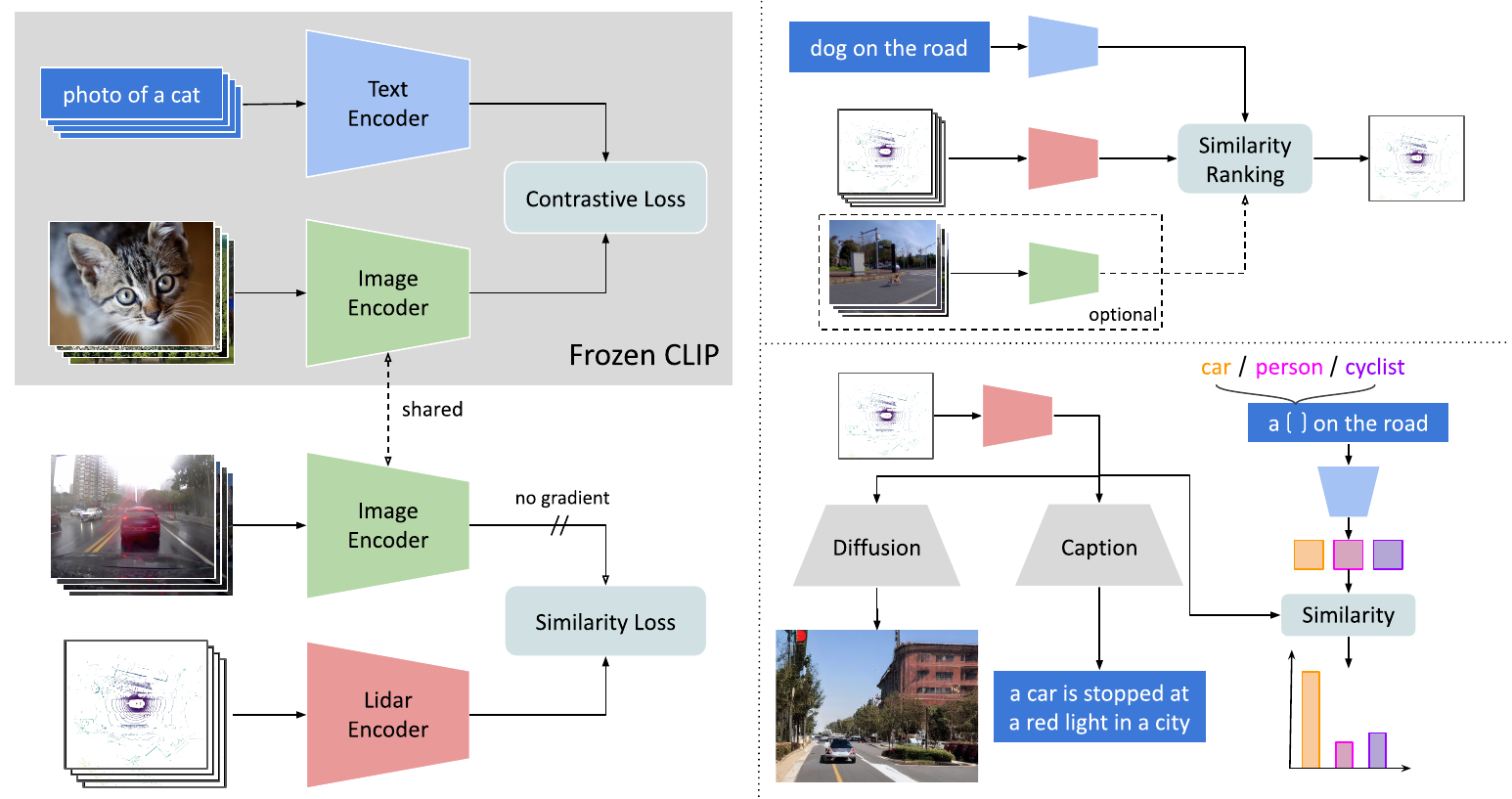}
    \caption{Overview of LidarCLIP. We use existing CLIP image and text encoders (top left), and learn to embed point clouds into the same feature space (bottom left). To that end, we train a lidar encoder to match the features of the frozen image encoder on a large automotive dataset with image-lidar pairs. This enables a wide range of applications, such as scenario retrieval (top right), zero-shot classification, as well as lidar-to-text and lidar-to-image generation (bottom right).}
    \label{fig:overview}
\end{figure*}

While conceptually simple, we demonstrate LidarCLIP's fine-grained semantic understanding for a wide range of applications. LidarCLIP outperforms prior works applying CLIP in the point cloud domain \cite{huang2022clip2point,zhang2022pointclip} on both zero-shot classification and retrieval. Furthermore, we demonstrate that LidarCLIP can be combined with regular CLIP to perform targeted searches for rare and difficult traffic scenarios, \eg, a person crossing the road while hidden by water drops on the camera lens, see \cref{fig:first_page}. Finally, LidarCLIP's capabilities are extended to point cloud captioning and lidar-to-image generation using established CLIP-based methods~\cite{clipcap,dhariwal2021diffusion}.

In summary, our contributions are the following:
\begin{itemize}
    \item We propose LidarCLIP, a new method for embedding lidar point clouds into an existing CLIP space.
    \item We demonstrate the effectiveness of LidarCLIP for retrieval and zero-shot classification, where it outperforms existing CLIP-based methods.
    \item We show that LidarCLIP is complementary to its CLIP teacher and even outperforms it in certain retrieval categories. By combining both methods, we further improve performance and enable retrieval of safety-critical scenes in challenging sensing conditions.
    \item Finally, we show that our approach enables a multitude of applications off-the-shelf, such as point cloud captioning and lidar-to-image generation.
\end{itemize}

\section{Related work}
\label{sec:related_work}
\noindent\textbf{CLIP and its applications.}
CLIP \cite{CLIP} is a model with a joint embedding space for images and text. The model consists of two encoders, a text encoder $\mathcal{F}_T$ and an image encoder $\mathcal{F}_I$, both of which yield a single feature vector describing their input. Using contrastive learning, these feature vectors have been supervised to map to a common language-visual space where images and text are similar if they describe the same scene. By training on 400 million text-image pairs collected from the internet, the model has a diverse textual understanding.

The shared text-image space can be used for many tasks. For instance, to do zero-shot classification with $K$ classes, one constructs $K$ text prompts, \eg, “a photo of a $\langle \text{class name}\rangle$”. These are embedded individually by the text encoder, yielding a feature map $Z_T\in \mathbb{R}^{K\times d}$. The logits for an image $I$ are calculated by comparing the image embedding, $\mathbf{z}_I\in\mathbb{R}^{d}$, with the feature map for the text prompts, $Z_T$,
and class probabilities $p$ are found using the softmax function, $\text{softmax}( Z_T \mathbf{z}_I)$. In theory, any concept encountered in the millions of text-image pairs could be classified with this approach. Further, by comparing a single prompt to multiple images, CLIP can also be used for retrieving images from a database.

Multiple works have built upon the CLIP embeddings for various applications. DALL-E 2 \cite{dalle2} and Stable Diffusion \cite{rombach2022high} are two methods that use the CLIP space for conditioning diffusion models for text-to-image generation. Other works have recently shown how to use text-image embeddings to generate single 3D objects \cite{Sanghi_2022_CVPR} and neural radiance fields \cite{Wang_2022_CVPR} from text. In \cite{zhou2022extract}, CLIP is used for zero-shot semantic segmentation without any labels. Similarly, \cite{wang2022cris} extracts pixel-level information for referring semantic segmentation, \ie, segmenting the part of an image referred to via a natural linguistic expression. We hope that LidarCLIP can spur similar applications for 3D data.

\noindent\textbf{CLIP outside the language-image domain.}
Besides new applications, multiple works have aimed to extend CLIP to new domains, and achieved impressive performance in their respective domains. For videos, CLIP has been used for tasks like video clip retrieval \cite{luo2022clip4clip,ma2022x} and video question answering \cite{xu-etal-2021-videoclip}. In contrast to our work, these methods rely on large amounts of text-video pairs for training. Meanwhile, WAV2CLIP \cite{wu2022wav2clip} and AudioCLIP \cite{9747631} extend CLIP to audio data for audio classification, tagging, and retrieval. Both methods use contrastive learning, which typically requires large batch sizes for convergence \cite{chen2020simple}. The scale of automotive point clouds would require extensive computational resources for contrastive learning, hence we supervise LidarCLIP with a simple mean squared error, which works well for smaller batch sizes and has been shown to promote the learning of richer features \cite{carlssoncross}. 

\noindent\textbf{Point clouds and natural language.}
Recently, there has been increasing interest in connecting point clouds and natural language, as it enables an intuitive interface for the 3D domain and opens up possibilities for open-vocabulary zero-shot learning. In \cite{cheraghian2022zero} and \cite{michele2021generative}, classifiers are supervised with pre-trained word embeddings to enable zero-shot learning. Part2word \cite{tang2021part2word} explores 3d shape retrieval by mapping scans of single objects and descriptive texts to a joint embedding space. However, a key limitation of these approaches is their need for dense annotations \cite{cheraghian2022zero,michele2021generative} or detailed textual descriptions \cite{tang2021part2word}, which makes them unable to leverage the vast amount of raw automotive data considered in this paper. 

Other methods, such as PointCLIP \cite{zhang2022pointclip} and CLIP2Point \cite{huang2022clip2point}, use CLIP to bypass the need for text-lidar pairs entirely. Instead of processing the point clouds directly, they render the point cloud from multiple viewpoints and apply the image encoder to these renderings. While this works well with dense point clouds of a single object, the approach is not feasible for sparse automotive data with heavy occlusions. In contrast, our method relies on an encoder specifically designed for the point cloud domain, avoiding the overhead introduced by multiple renderings and allowing for more flexibility in the model choice.

\section{LidarCLIP}
\label{sec:method}
In this work, we encode lidar point clouds into the existing CLIP embedding space. As there are no datasets with text-lidar pairs, we cannot rely on the same contrastive learning strategy as the original CLIP model to directly relate point clouds to text. Instead, we leverage that automotive datasets contain millions of image-lidar pairs. By training a point cloud encoder to mimic the features of a frozen CLIP image encoder, the images act as intermediaries for connecting text and point clouds, see \cref{fig:overview}.

Each training pair consists of an image $\mathbf{x}_I$ and the corresponding point cloud $\mathbf{x}_L$. Regular CLIP does not perform alignment between the pairs, but some preprocessing is needed for point clouds. To align the contents of both modalities, we transform the point cloud to the camera coordinate system and drop all points that are not visible in the image. As a consequence, we only perform inference on frustums of the point cloud, corresponding to a typical camera field of view. We note that this preprocessing is susceptible to errors in sensor calibration and time synchronization, especially for objects along the edge of the field of view. Further, the preprocessing does not handle differences in visibility due to sensor mounting positions, \eg, lidars are typically mounted higher than cameras in data collection vehicles, thus seeing over some vehicles or static objects. However, using millions of training pairs reduces the impact from such noise sources. 

The training itself is straightforward. An image is passed through the frozen image encoder $\mathcal{F}_I$ to produce the target embedding,
\begin{align}
    \mathbf{z}_I = \mathcal{F}_I(\mathbf{x}_I),
\end{align}
whereas the lidar encoder $\mathcal{F}_L$ embeds a point cloud,
\begin{align}
    \mathbf{z}_L = \mathcal{F}_L(\mathbf{x}_L).
\end{align}
We train $\mathcal{F}_L$ to maximize the similarity between $\mathbf{z}_I$ and $\mathbf{z}_L$. To this end, we adopt either the mean squared error (MSE),
\begin{align}
    \mathcal{L}_\text{MSE} = \frac{1}{d}(\mathbf{z}_I - \mathbf{z}_L)^T(\mathbf{z}_I - \mathbf{z}_L),
\end{align}
or the cosine similarity loss,
\begin{align}
    \mathcal{L}_\text{cos} = -\frac{\mathbf{z}_I^T \mathbf{z}_L}{\Vert\mathbf{z}_I\Vert \Vert\mathbf{z}_L\Vert},
\end{align}
where $\mathbf{z}_I, \mathbf{z}_L \in \mathbb{R}^d$.
The main advantage of using a similarity loss that only considers positive pairs, as opposed to using a contrastive loss, is that we avoid the need for large batch sizes \cite{chen2020simple} and the accompanying computational requirements. Furthermore, the benefits of contrastive learning are reduced in our settings, since we only care about mapping a new modality into an existing feature space, rather than learning an expressive feature space from scratch. 

\subsection{Joint retrieval}
\label{sec:joint-retrieval}
Retrieval is one of the most successful applications of CLIP and is highly relevant for the automotive industry. By retrieval, we mean the process of finding samples that best match a given natural language prompt out of all the samples in a large database. In an automotive setting, it is used to sift through the abundant raw data for valuable samples. While CLIP works well for retrieval out of the box, it inherits the fundamental limitations of the camera modality, such as poor performance in darkness, glare, or water spray. LidarCLIP can increase robustness by leveraging the complementary properties of lidar. 

Relevant samples are retrieved by computing the similarity between a text query and each sample in the database, in the CLIP embedding space, and identifying the samples with the highest similarity. These calculations may seem expensive, but the embeddings only need to be computed once per sample, after which they can be cached and reused for every text query. Following prior work \cite{CLIP}, we compute the retrieval score using cosine similarity for both image and lidar
\begin{equation}
    s_I = \frac{\mathbf{z}_T^T \mathbf{z}_I}{\Vert\mathbf{z}_T\Vert \Vert\mathbf{z}_I\Vert}, \quad \quad s_L = \frac{\mathbf{z}_T^T \mathbf{z}_L}{{\Vert\mathbf{z}_T\Vert \Vert\mathbf{z}_L\Vert}},
\end{equation}
where $\mathbf{z}_T$ is the text embedding. If a database only contains images or point clouds, we use the corresponding score ($s_I$ or $s_L$) for retrieval. However, if we have access to both images and point clouds, we can jointly consider the lidar and image embeddings to leverage their respective strengths.

We consider various methods of performing joint retrieval. Inspired by the literature on ensembles \cite{Ganaie_2022}, we can directly combine the image and lidar similarity scores, $s_{I+L} = s_L + s_I$. We can also fuse the modalities even earlier by combining the features, $\mathbf{z}_{I+L} = \mathbf{z}_I + \mathbf{z}_L$. We also consider methods to aggregate independent rankings for each modality. One such approach is to consider the joint rank to be the mean rank across the modalities. Another approach we have evaluated is two-step re-ranking \cite{Min2021JointPR}, where one modality selects a set of candidates which are then ranked by the other modality.

One of the most exciting aspects of joint retrieval is the possibility of using different queries for each modality. For example, imagine trying to find scenes where a large white truck is almost invisible in the image due to extreme sun glare. In this case, one can search for scenes where the image embedding matches ``an image with extreme sun glare'' and re-rank the top-K results by their lidar embeddings similarity to ``a scene containing a large truck''. This kind of scene would be almost impossible to retrieve using a single modality.

\section{Experiments}
\label{sec:results}

\noindent\textbf{Datasets.} Training and most of the evaluation is done on the large-scale ONCE dataset \cite{mao2021one}, with roughly 1 million scenes. Each scene consists of a lidar sweep and 7 camera images, which results in ${\raise.17ex\hbox{$\scriptstyle\sim$}}7$ million unique training pairs. We withhold the validation and test sets and use these for the results presented below.

\noindent\textbf{Implementation details.} We use the official CLIP package and models, specifically the most capable vision encoder, ViT-L/14, which has a feature dimension $d=768$. As our lidar encoder, we use the Single-stride Sparse Transformer (SST) \cite{Fan_2022_CVPR} (randomly initialized). Due to computational constraints, our version of SST is down-scaled and contains about 8.5M parameters, which can be compared to the ${\raise.17ex\hbox{$\scriptstyle\sim$}}85$M and ${\raise.17ex\hbox{$\scriptstyle\sim$}}300$M parameters of the text and vision encoders of CLIP. The specific choice of backbone is not key to our approach; similar to the variety of CLIP image encoders, one could use a variety of different lidar encoders. However, we choose a transformer-based encoder, inspired by the findings that CLIP transformers perform better than CLIP ResNets \cite{CLIP}. SST is trained for 3 epochs, corresponding to ${\raise.17ex\hbox{$\scriptstyle\sim$}}20$ million training examples, using the Adam optimizer and the one-cycle learning rate policy. For full details, we refer to our code.

\noindent\textbf{Retrieval ground truth \& prompts.} One difficulty in quantitatively evaluating the retrieval capabilities of LidarCLIP is the lack of direct ground truth for the task. Instead, automotive datasets typically have fine-grained annotations for each scene, such as object bounding boxes, segmentation masks, etc. This is also true for ONCE, which contains annotations in terms of 2D and 3D bounding boxes for 5 classes, and metadata for the time of day and weather. We leverage these detailed annotations and available metadata, to create as many retrieval categories as possible. For object retrieval, we consider a scene positive if it contains one or more instances of that object. To probe the spatial understanding of the model, we also propose a ``nearby'' category, searching specifically for objects closer than \SI{15}{\meter}. We verify that the conclusions hold for thresholds between \SI{10}{\meter} and \SI{25}{\meter}. Finally, to minimize the effect of prompt engineering, we follow \cite{Gu2022OpenvocabularyOD} and average multiple text embeddings to improve results and reduce variability. For object retrieval, we use the same 85 prompt templates as in \cite{Gu2022OpenvocabularyOD}, and for the other retrieval categories, we use similar patterns to generate numerous relevant prompts templates. The exact prompts are provided in the source code.

\subsection{Zero-shot classification}
\label{sec:zs-cls}

\begin{table}[t]
    \centering
    \label{tab:zscls}
    \begin{tabular}{lll}
    \toprule
                                             & Cls.                & Obj.               \\ \midrule
        PointCLIP \cite{zhang2022pointclip}   & $23.8\%$            & $34.3\%$          \\
        CLIP2Point \cite{huang2022clip2point} & $32.1\%$            & $26.2\%$          \\
        CLIP2Point w/ pre-train \cite{huang2022clip2point} & $29.8\%$ & $28.2\%$        \\
        LidarCLIP                            & $\mathbf{43.6\%}$   & $\mathbf{62.1\%}$  \\ \bottomrule
    \end{tabular}
    \caption{Zero-shot classification on ONCE \textit{val}, top-1 accuracy averaged over classes/objects.}
\end{table}

CLIP's \cite{CLIP} strong open-vocabulary capabilities have made it popular for zero-shot classification. We construct this task by treating each annotated object in ONCE as a separate classification sample. Typically, LidarCLIP outputs a set of voxel features that are pooled into a single, global, CLIP feature. We construct the object embeddings by only pooling features for voxels inside the corresponding bounding box, without any object-specific training/fine-tuning. We compare our performance to PointCLIP \cite{zhang2022pointclip} and CLIP2Point \cite{huang2022clip2point}, two works that transfer CLIP to 3d by rendering point clouds from multiple viewpoints and then apply the CLIP image encoder and pool the features from different views. For CLIP2Point, we also include their provided weights, which have been pre-trained on ModelNet40. We create their object embeddings by extracting only points within the annotated bounding boxes and adapt their standard evaluation protocol. Results for LidarCLIP and their best-performing model are shown in \cref{tab:zscls} where we report top-1 accuracy average over objects and classes, as the data contains a few majority classes. The results demonstrate the gain from training a modality specific encoder rather than transferring point clouds into the image domain. Further, we should note that LidarCLIP's instance-level classification performance is achieved without any dense annotations in 3d or 2d. Applications of LidarCLIP for zero-shot point cloud semantic segmentation is left for future work.

\subsection{Retrieval}

\begin{table*}
    \centering
        \begin{tabular}{@{}lllllllllllllll@{}}
        \toprule
        \multicolumn{1}{l|}{P@K}   & 10           & 100           & 10           & 100           & 10           & 100           & 10           & 100           & 10           & \multicolumn{1}{l|}{100}            & 10            & 100           & \multicolumn{1}{|l}{10} & 100\\ \midrule
        \multicolumn{1}{l|}{Object Cat.}       & \multicolumn{2}{c}{Car}      & \multicolumn{2}{c}{Truck}    & \multicolumn{2}{c}{Bus}      & \multicolumn{2}{c}{Pedestrian}   & \multicolumn{2}{c}{Cyclist}                    & \multicolumn{2}{|c}{Avg.}     & \multicolumn{2}{|c}{Avg. Nearby}    \\ \midrule
        \multicolumn{1}{l|}{Image} & 1.0          & 0.96          & 0.9          & \textbf{0.94} & 0.9          & 0.94          & 0.9          & 0.83          & 0.5          & \multicolumn{1}{l|}{0.37}          & 0.84          & 0.81           & \multicolumn{1}{|l}{0.76} & 0.67\\
        \multicolumn{1}{l|}{Lidar} & \textbf{1.0} & \textbf{0.99} & 0.8          & 0.74          & \textbf{1.0} & \textbf{0.97} & 0.8          & 0.82          & \textbf{1.0} & \multicolumn{1}{l|}{\textbf{0.58}} & 0.92          & 0.82           & \multicolumn{1}{|l}{0.88} & 0.78\\
        \multicolumn{1}{l|}{Joint} & 0.9          & 0.97          & \textbf{1.0} & 0.92          & \textbf{1.0} & \textbf{0.97} & \textbf{1.0} & \textbf{0.90} & 0.9          & \multicolumn{1}{l|}{0.42}          & \textbf{0.96} & \textbf{0.84}  & \multicolumn{1}{|l}{\textbf{0.90}} & \textbf{0.81}\\ \bottomrule
        \end{tabular}
    \caption{Retrieval for scenes containing various object categories. We report precision at ranks 10 and 100. Notice the joint classification is superior overall, but there are two categories (bus and cycle), where using only lidar is advantageous.}
    \label{tab:retrieval-objects}
\end{table*}

\begin{table*}
    \centering
        \begin{tabular}{lllllllllllllllllllll}
        \toprule
        \multicolumn{1}{l|}{P@K}    & 10           & 100           & 10           & 100           & 10           & 100           & 10           & 100           & 10           & 100           & 10           & 100 & \multicolumn{1}{|l}{10} & 100          \\ \midrule
        \multicolumn{1}{l|}{Scene Cat.}        & \multicolumn{2}{c}{Night}    & \multicolumn{2}{c}{Day}      & \multicolumn{2}{c}{Sunny}    & \multicolumn{2}{c}{Rainy}    & \multicolumn{2}{c}{Busy}     & \multicolumn{2}{c}{Empty} & \multicolumn{2}{|c}{Avg.}      \\ \midrule
        \multicolumn{1}{l|}{Random} & \multicolumn{2}{c}{0.25}     & \multicolumn{2}{c}{0.75}     & \multicolumn{2}{c}{0.43}     & \multicolumn{2}{c}{0.50}     & \multicolumn{2}{c}{0.23}     & \multicolumn{2}{c}{0.35}  & \multicolumn{2}{|c}{0.42}        \\
        \multicolumn{1}{l|}{Image}  & \textbf{1.0} & \textbf{1.00} & 0.9          & 0.92          & \textbf{1.0} & \textbf{1.00} & \textbf{1.0} & \textbf{1.00} & \textbf{1.0} & 0.91          & \textbf{0.8} & 0.73 & \multicolumn{1}{|l}{0.95} & 0.93         \\
        \multicolumn{1}{l|}{Lidar}  & 0.2          & 0.53          & \textbf{1.0} & 0.98          & 0.5          & 0.74          & 0.8          & 0.98          & \textbf{1.0} & 0.93          & 0.5          & 0.69   & \multicolumn{1}{|l}{0.67} & 0.81       \\
        \multicolumn{1}{l|}{Joint}  & \textbf{1.0} & \textbf{1.00} & \textbf{1.0} & \textbf{0.99} & \textbf{1.0} & \textbf{1.00} & \textbf{1.0} & \textbf{1.00} & \textbf{1.0} & \textbf{0.99} & \textbf{0.8} & \textbf{0.79} & \multicolumn{1}{|l}{\textbf{0.97}} & \textbf{0.96} \\ \bottomrule
        \end{tabular}
    \caption{Retrieval of scenes with various global conditions. We report precision at ranks 10 and 100. Notice that the joint classification is superior for all types of conditions. }
    \label{tab:retrieval-scene-conditions}
\end{table*}

To evaluate retrieval, we report the commonly used Precision at rank K (P@K) \cite{ghorab2013personalised,ma2022ei,rehman2012content}, for $K = {10,100}$, which measures the fraction of positive samples within the top K predictions. Recall at K is another commonly used metric \cite{ghorab2013personalised,rehman2012content}, however, it is hard to interpret when the number of positives is in the thousands, as is the case here. We evaluate the performance of three approaches: lidar-only, camera-only, and the joint approach proposed in \cref{sec:joint-retrieval}. We perform retrieval for scenes containing various kinds of objects and present the results in \cref{tab:retrieval-objects}. We also evaluate PointCLIP \cite{zhang2022pointclip} and CLIP2Point \cite{huang2022clip2point}, however, their methods are not suited for the large-scale point clouds considered here and consequently barely outperform random guessing.

\noindent\textbf{Object-level.} Interestingly, lidar retrieval performs slightly better than image retrieval on average, despite being trained to mimic the image features, with the biggest improvements in the cyclist class. A possible explanation is that cyclist features are more invariant in the point cloud than in the image, allowing the lidar encoder to generalize to cyclists that go undetected by the image encoder. A noteworthy problem for lidar retrieval is trucks. Upon qualitative inspection, we find that the lidar encoder confuses trucks with buses, which is a drawback with certain objects' features being more invariant in lidar data. We also attempt to retrieve scenes where objects of the given class appear close to the ego vehicle. Here, we can see that joint retrieval truly shines, greatly outperforming single-modality retrieval in classes like truck and pedestrian. One interpretation is that the lidar is more reliable at determining distance, while the image can be leveraged to distinguish between classes (such as trucks and buses) based on textures and other fine details only visible in the image.

\noindent\textbf{Scene-level.} Object-centric retrieval is focused on \textit{local} details of a scene and should trigger even for a single occluded pedestrian on the side of the road. Therefore, we run another set of experiments focusing on \textit{global} properties such as weather, time of day, and general `busyness' of the scene. In \cref{tab:retrieval-scene-conditions}, we see that the lidar is outperformed by the camera for determining light conditions. This seems quite expected, and if anything, it is somewhat surprising that lidar can do significantly better than random in these categories. Again, we see that joint retrieval consistently gets the best of both worlds and, in some cases, such as empty scenes, clearly outperforms both single-modality methods.

\begin{figure}[t]
    \centering
    \begin{tabular}{cc}
        \includegraphics[width=0.45\linewidth]{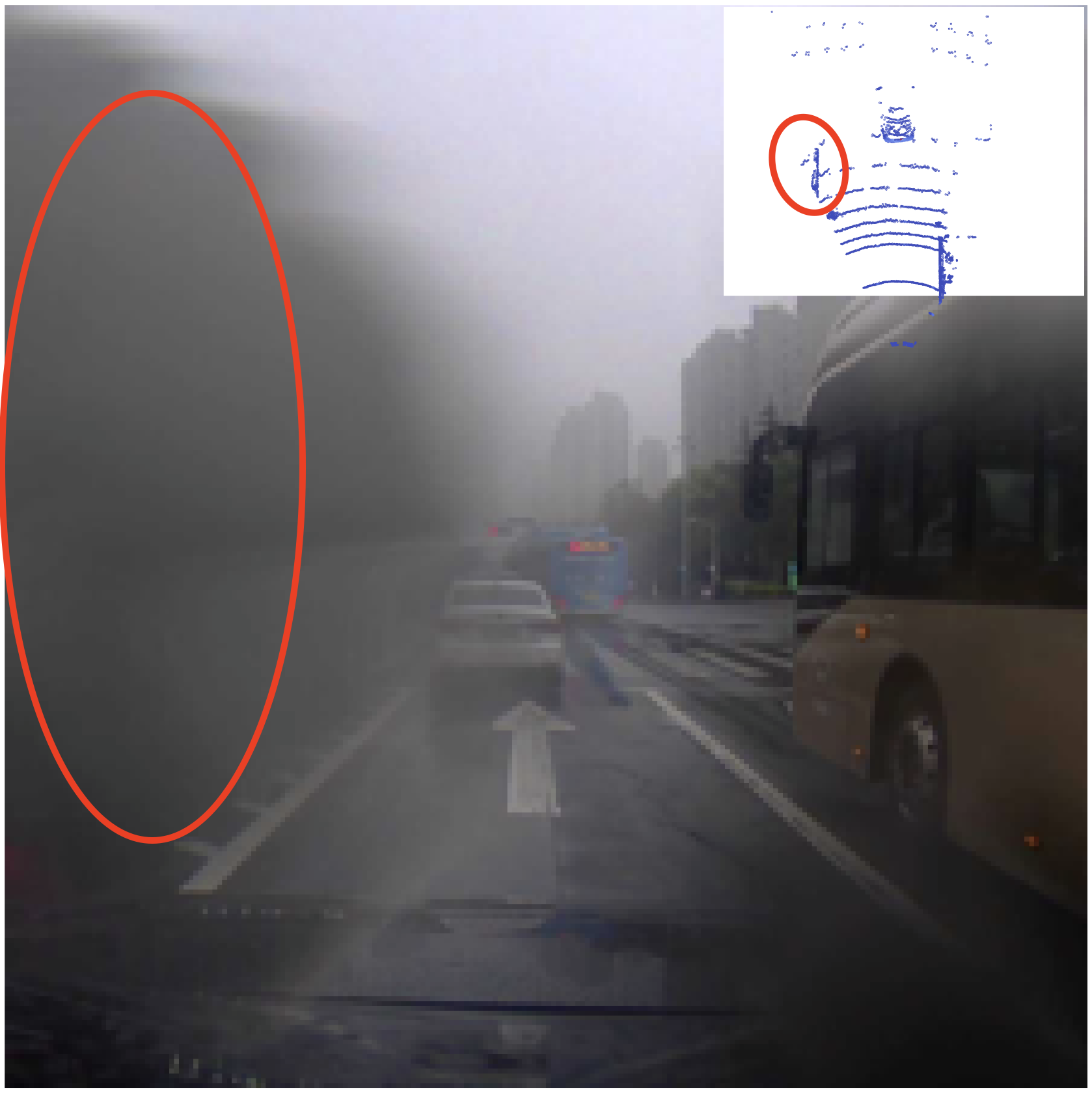} &
        \includegraphics[width=0.45\linewidth]{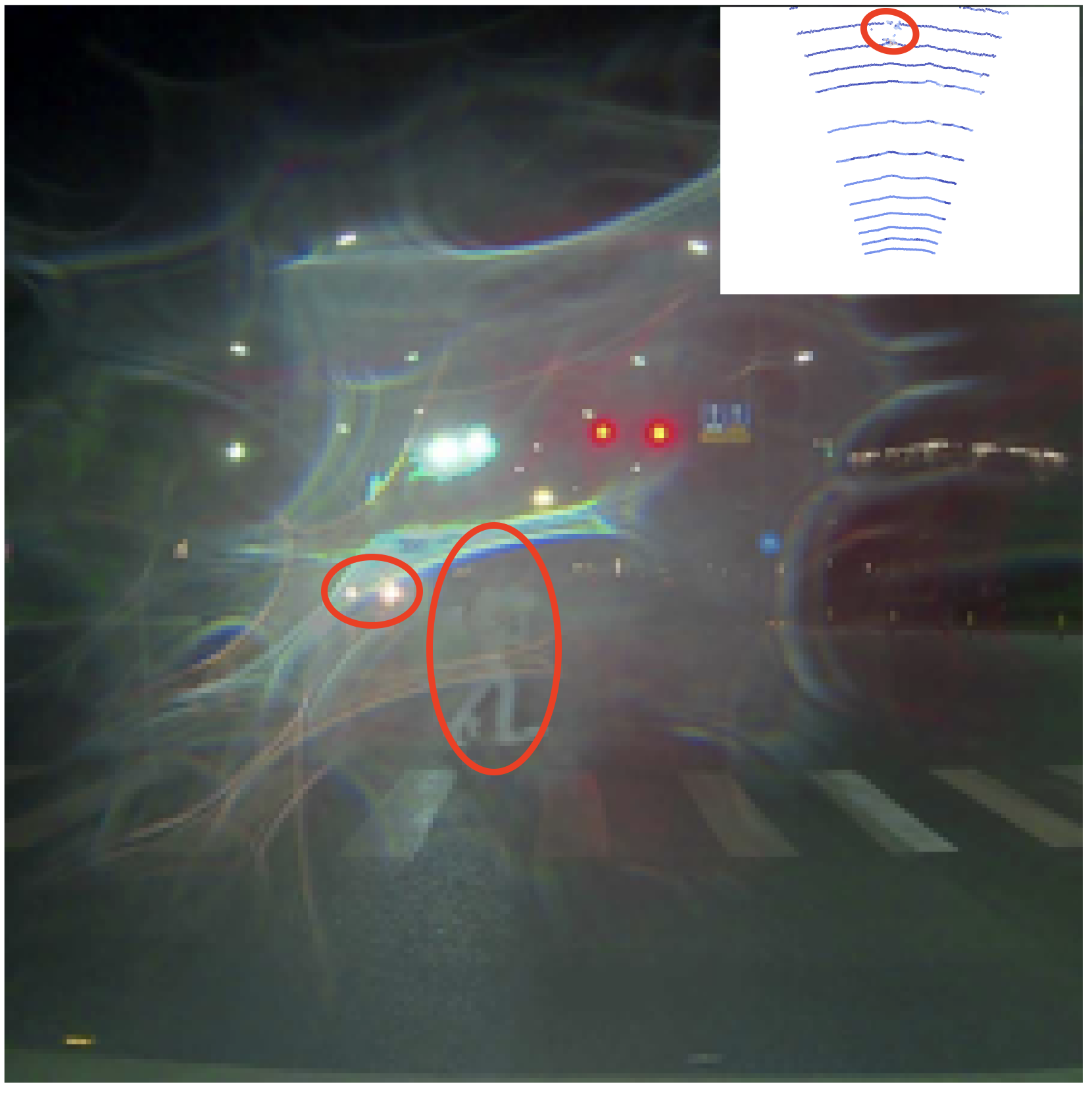} 
    \end{tabular}
    \caption{Example of retrieval using separate prompts for image and lidar. We query for images with blur, water spray, glare, corruption, and lack of objects, and for point clouds with nearby trucks, pedestrians, cars, etc. By combining the scores of these separate queries, we can find edge cases that are extremely valuable during the training/validation of a camera-based perception system. These valuable objects are highlighted in red, both in the image and point cloud.}
    \label{fig:joint_blur}

\end{figure}

\noindent\textbf{Separate prompts.} Inspired by the success of joint retrieval, and the complementary aspects of sensing in lidar and camera, we present some qualitative examples where different prompts are used for each modality. Thus, we can find scenes that are difficult to identify with a single modality. \cref{fig:joint_blur} shows retrieval examples where the image was prompted for glare, extreme blur, water on the lens, corruption, and lack of objects in the scene. At the same time, the lidar was prompted for nearby objects such as cars, trucks, and pedestrians. As seen in \cref{fig:joint_blur}, the examples indicate that we can retrieve scenes where these objects are almost completely invisible in the image. Such samples are highly valuable both for the training and validation of autonomous driving systems.

\begin{table}[t]
    \centering
        \begin{tabular}{l|ll}
        \toprule
        \multicolumn{1}{c|}{Loss function } & \multicolumn{1}{c}{P@10} & \multicolumn{1}{c}{P@100} \\ \midrule
        Mean squared error    & \textbf{0.869}           & \textbf{0.810}            \\
        Cosine similarity     & 0.781                    & 0.748                     \\ \midrule
        \end{tabular}
    \caption{Ablation of the LidarCLIP training loss. We report precision at ranks 10 and 100, averaged over all prompts. Training with MSE leads to better retrieval performance.}
    \label{tab:loss-ablation}
\end{table}

\begin{table}[t]
    \centering
        \begin{tabular}{l|ll}
        \toprule
        \multicolumn{1}{l|}{Method} & \multicolumn{1}{c}{P@10}           & \multicolumn{1}{c}{P@100} \\ \midrule
        Image only                  & 0.856          & 0.810                     \\
        Lidar only                  & 0.813          & 0.803                     \\ \midrule
        Mean feature                & \textbf{0.944} & \textbf{0.876}            \\
        Mean norm. feature          & \textbf{0.944} & 0.875                     \\
        Mean score                  & 0.919          & 0.874                     \\
        Mean ranking                & 0.888          & 0.854                     \\
        Reranking - img first       & 0.925          & 0.867                     \\
        Reranking - lidar first     & 0.875          & 0.860                     \\ \bottomrule
        \end{tabular}
    \caption{Ablation of joint retrieval methods. We report precision at ranks 10 and 100, averaged over all prompts. All methods improve upon the single-modality models, but averaging lidar and image features before normalization achieves the best performance.}
    \label{tab:joint-retrieval-ablation}
\end{table}

\noindent\textbf{Domain transfer.} For studying the robustness of LidarCLIP under domain shift, we evaluate its retrieval performance on a different dataset than it was trained on, namely, the nuScenes dataset \cite{caesar2020nuscenes}. Compared to ONCE, the nuScenes lidar sensor has fewer beams (32 vs 40), lower horizontal resolution, and different intensity characteristics. Further, nuScenes is collected in Boston and Singapore, while ONCE is collected in Chinese cities. The challenge of transferring between these datasets has been shown in unsupervised domain adaptation \cite{mao2021one}. Similarly to the ONCE retrieval task, we generate ground truth using the object annotations.

We compare the model trained on ONCE with a reference model trained directly on nuScenes in \cref{tab:nuscenes}. As expected, the differences in sensor characteristics hamper the ability to perform lidar-only retrieval on the target dataset. Notably, we find that the joint method is robust to this effect, showing almost no domain transfer gap, and outperforming camera-only retrieval even with the ONCE-trained lidar encoder.

\begin{table}[]
    \centering
        \begin{tabular}{l|ccccll}
            \toprule
            Train set & \multicolumn{2}{c}{ONCE}  & \multicolumn{2}{c}{nuScenes} \\ \midrule
            P@K   & 10    & 100  & 10    & 100   \\
            \midrule
            Image & 0.69  & 0.65  & 0.69  & 0.65  \\
            Lidar & 0.46  & 0.40  & 0.79  & 0.64  \\
            Joint  & \textbf{0.74}  &\textbf{0.69}  & \textbf{0.81}  & \textbf{0.70}  \\ \bottomrule
        \end{tabular}
    \caption{nuScenes \textit{val} retrieval with different \textit{train} sets. Performance is averaged over classes. LidarCLIP supports the joint retrieval, even when trained and evaluated on separate datasets.}
    \label{tab:nuscenes}
\end{table}

\noindent\textbf{Ablations.} As described in \cref{sec:method}, we have two primary candidates for the training loss function. MSE encourages the model to embed the point cloud in the same position as the image, whereas cosine similarity only cares about matching the directions of the two embeddings. We compare the retrieval performance of two models trained using these losses in \cref{tab:loss-ablation}. To reduce training time, we use the ViT-B/32 CLIP version, rather than the heavier ViT-L/14. The results show that using MSE leads to significantly better retrieval, even though retrieval uses cosine similarity as the scoring function. We also perform ablations on the different approaches for joint retrieval described in \cref{sec:joint-retrieval}. As shown in \cref{tab:joint-retrieval-ablation}, the simple approach of averaging the camera and lidar features gives the best performance, and it is thus the approach used throughout the paper.

\noindent\textbf{Investigating lidar sensing capabilities.} Besides its usefulness for retrieval, LidarCLIP can offer more understanding of what concepts can be captured with a lidar sensor. While lidar data is often used in tasks such as object detection \cite{9181591}, panoptic/semantic segmentation \cite{aygun20214d,jhaldiyal2022semantic}, and localization \cite{9304812}, research into capturing more abstract concepts with lidar data is limited and focused mainly on weather classification~\cite{heinzler2019weather,s20154306}. However, we show that LidarCLIP can indeed capture complex scene concepts, as already demonstrated in \cref{tab:retrieval-scene-conditions}. 

\begin{figure}[t]
    \begin{tabular}{ccccc}
        \multicolumn{5}{c}{''a white car''} \\
        \includegraphics[width=0.19 \linewidth]{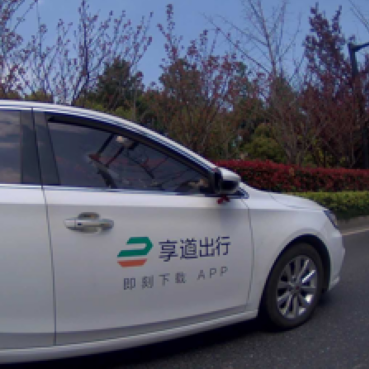} & \includegraphics[width=0.19 \linewidth]{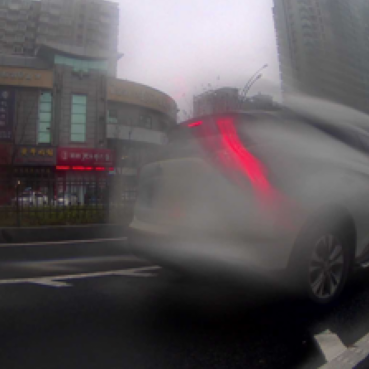} & \includegraphics[width=0.19 \linewidth]{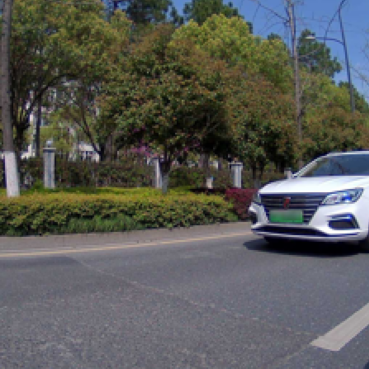} & \includegraphics[width=0.19 \linewidth]{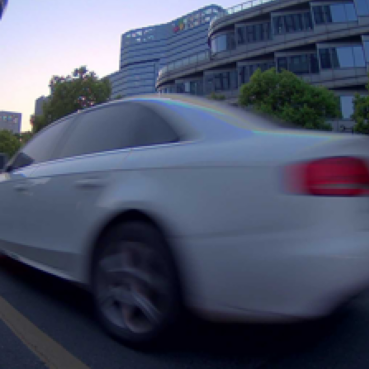} & \includegraphics[width=0.19 \linewidth]{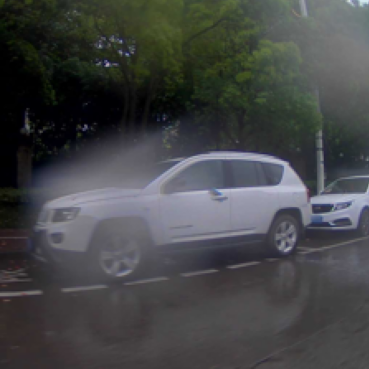} \\
        \multicolumn{5}{c}{''a black car''} \\
        \includegraphics[width=0.19 \linewidth]{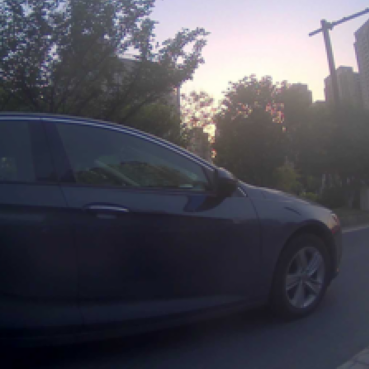} & \includegraphics[width=0.19 \linewidth]{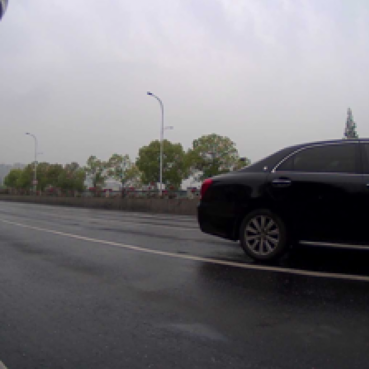} & \includegraphics[width=0.19 \linewidth]{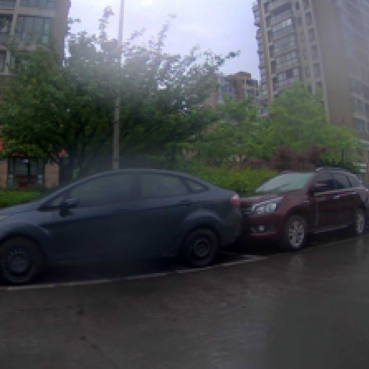} & \includegraphics[width=0.19 \linewidth]{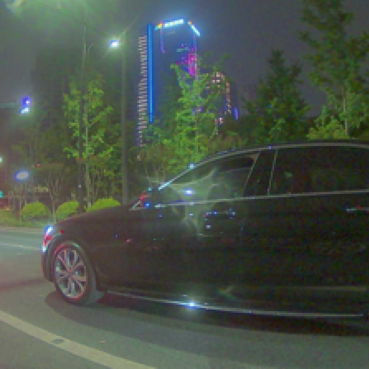} & \includegraphics[width=0.19 \linewidth]{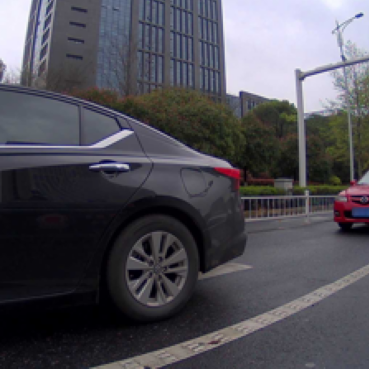} \\
        \multicolumn{5}{c}{''a red car''} \\
        \includegraphics[width=0.19 \linewidth]{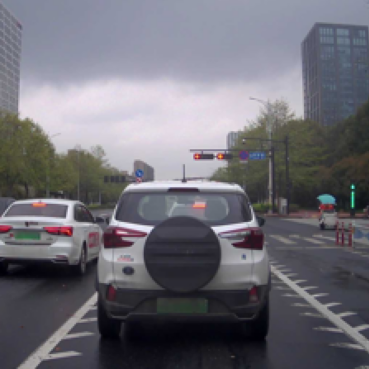} & \includegraphics[width=0.19 \linewidth]{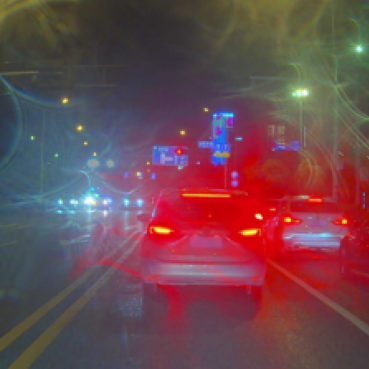} & \includegraphics[width=0.19 \linewidth]{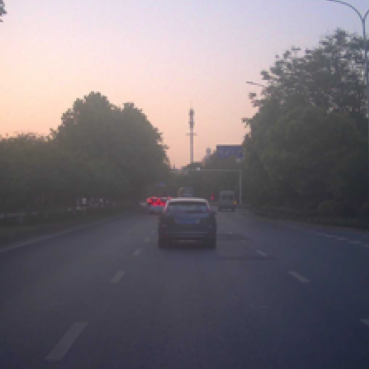} & \includegraphics[width=0.19 \linewidth]{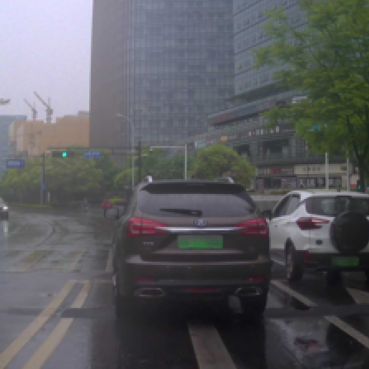} & \includegraphics[width=0.19 \linewidth]{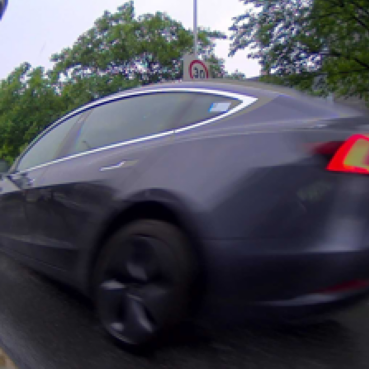}
    \end{tabular}
    \centering
    \caption{Top-5 retrieved examples from LidarCLIP for different colors. Note that images are only for visualization, point clouds were used for retrieval. LidarCLIP consistently differentiates black and white but struggles with specific colors.}
    \label{fig:color_retrieval}
\end{figure}

\begin{figure}[t]
    \begin{tabular}{ccccc}
        \multicolumn{5}{c}{bright and sunny} \\
        \includegraphics[width=0.19 \linewidth]{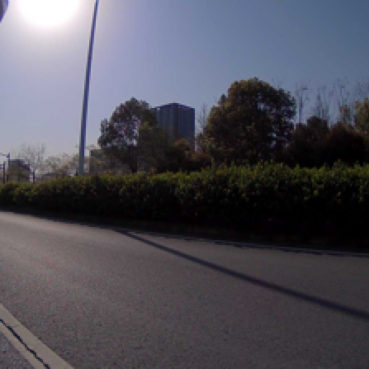} & \includegraphics[width=0.19 \linewidth]{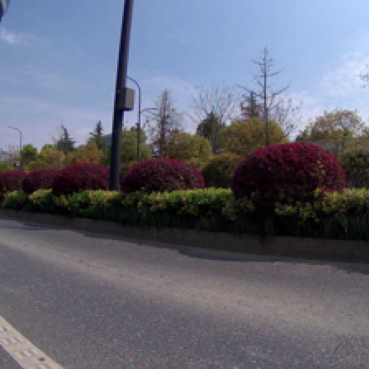} & \includegraphics[width=0.19 \linewidth]{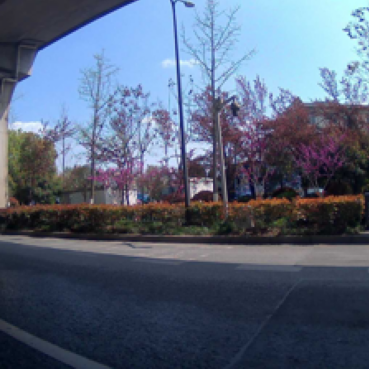} & \includegraphics[width=0.19 \linewidth]{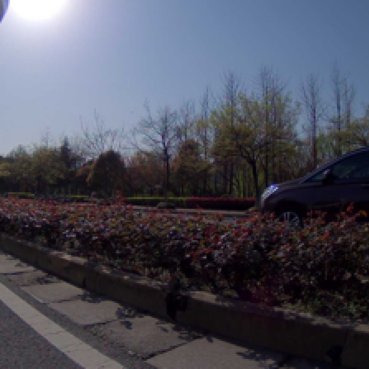} & \includegraphics[width=0.19 \linewidth]{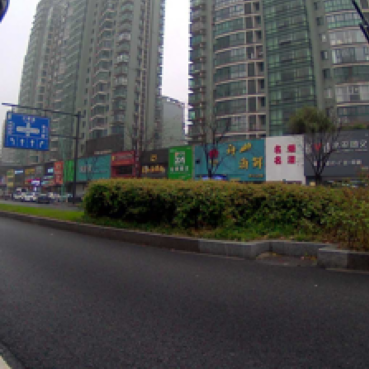} \\
        \multicolumn{5}{c}{sunrise} \\
        \includegraphics[width=0.19 \linewidth]{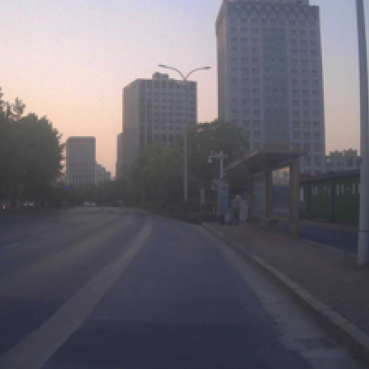} & \includegraphics[width=0.19 \linewidth]{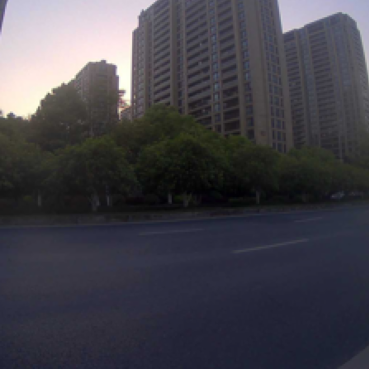} & \includegraphics[width=0.19 \linewidth]{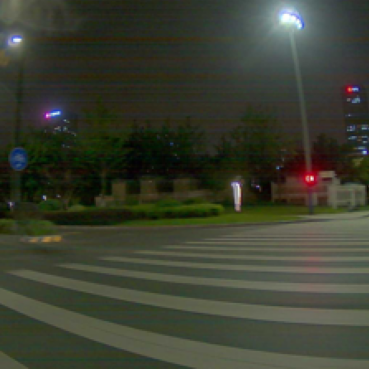} & \includegraphics[width=0.19 \linewidth]{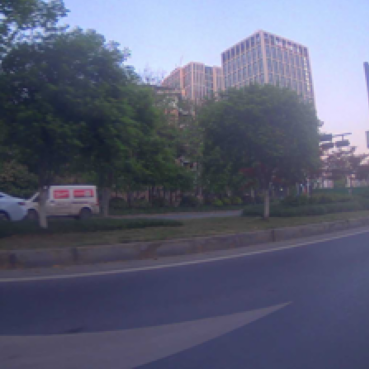} & \includegraphics[width=0.19 \linewidth]{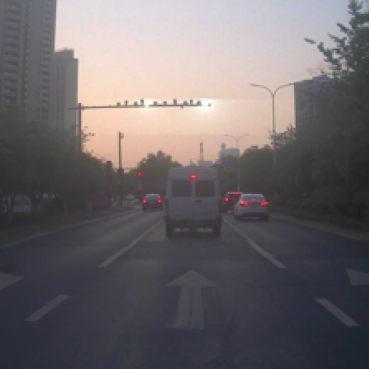} \\
        \multicolumn{5}{c}{night} \\
        \includegraphics[width=0.19 \linewidth]{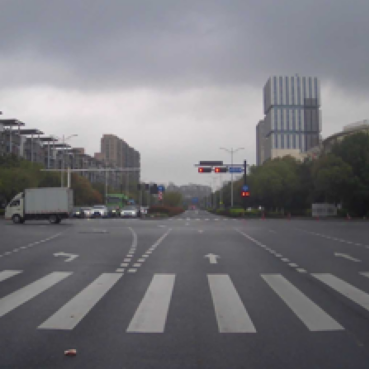} & \includegraphics[width=0.19 \linewidth]{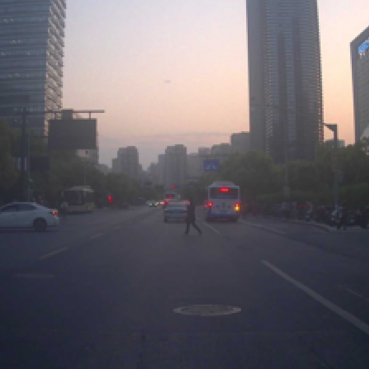} & \includegraphics[width=0.19 \linewidth]{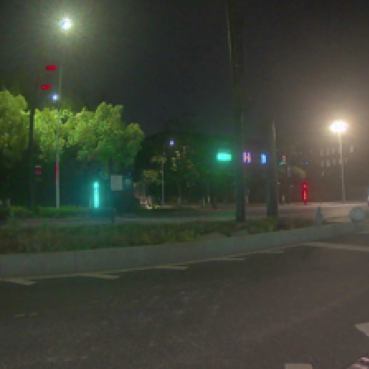} & \includegraphics[width=0.19 \linewidth]{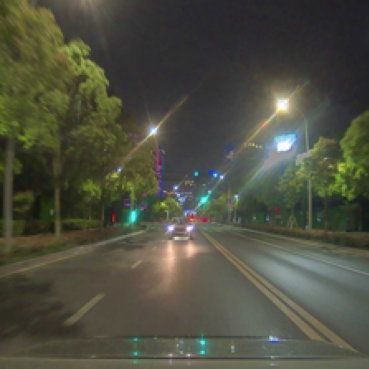} & \includegraphics[width=0.19 \linewidth]{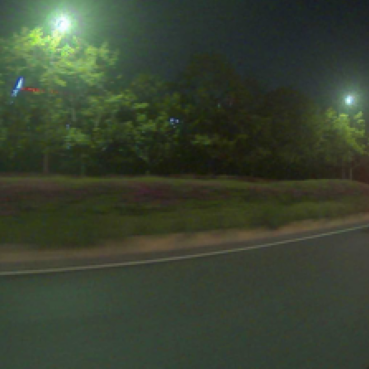} \\
        \multicolumn{5}{c}{headlights} \\
        \includegraphics[width=0.19 \linewidth]{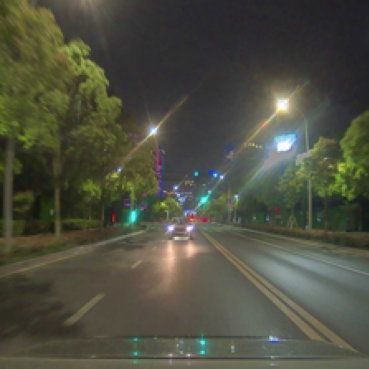} & \includegraphics[width=0.19 \linewidth]{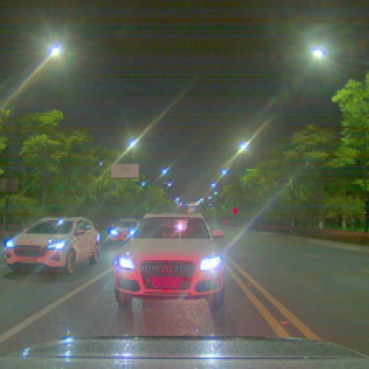} & \includegraphics[width=0.19 \linewidth]{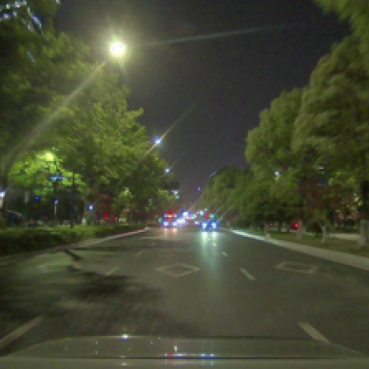} & \includegraphics[width=0.19 \linewidth]{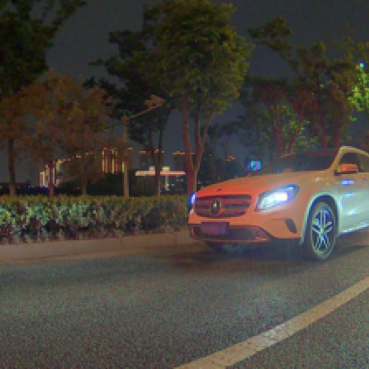} & \includegraphics[width=0.19 \linewidth]{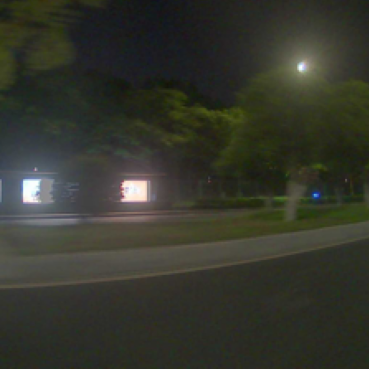}
    \end{tabular}
    \centering
    \caption{Top-5 retrieved examples from LidarCLIP for different lighting conditions (image only for visualization). LidarCLIP is surprisingly good at understanding the lighting of the scene, to the point of picking up on oncoming headlights with great accuracy.}
    \label{fig:lighting_retrieval}
\end{figure}

Inspired by this, we investigate the ability of LidarCLIP to extract color information, by retrieving scenes with “a $\langle \text{color}\rangle$~car”. As illustrated in Figure \ref{fig:color_retrieval}, while LidarCLIP struggles to capture specific colors accurately, it consistently differentiates between bright and dark colors. Such partial color information may be valuable for systems fusing lidar and camera information. Additionally, LidarCLIP learns meaningful features for overall scene lighting conditions, as illustrated in Figure \ref{fig:lighting_retrieval}. It can retrieve scenes based on the time of day, and is even able to distinguish scenes with many headlights from regular night scenes. Notably, all retrieved scenes are sparsely populated, indicating that LidarCLIP does not rely on biases associated with street congestion at different times of the day.

\setlength{\tabcolsep}{1pt}
\setlength{\tabcolsep}{6pt}

\setlength{\tabcolsep}{1pt}
\setlength{\tabcolsep}{6pt}


\subsection{Generative applications}
To demonstrate the flexibility of LidarCLIP, we integrate it with two existing CLIP-based generative models. For lidar-to-text generation, we utilize an image captioning model called ClipCap \cite{clipcap}, and for lidar-to-image generation, we use CLIP-guided Stable Diffusion\footnote{\url{https://github.com/huggingface/diffusers/tree/main/examples/community}}~\cite{rombach2022high}. In both cases, we replace the expected text or image embeddings with our point cloud embedding.

We evaluate image generation with the widely used Fréchet Inception Distance (FID) \cite{parmar2022aliased_short}. For this, we randomly select $\approx$6000 images from ONCE \textit{val} and generate images using CLIP-generated captions, CLIP features, or a combination of both. While FID is widely used, it has been shown to sometimes align poorly with human judgment \cite{kynkaanniemi2022role}. To complement this evaluation, we use CLIP-FID, with a different CLIP model to avoid any bias. We also implement pix2pix \cite{isola2017image} as a baseline for lidar-to-image generation. Notably, this setting not only evaluates the image generation performance but also serves as a proxy for assessing the captioning quality. Our results, presented in \cref{tab:image_generation}, demonstrate that incorporating captions significantly improves the photo-realism of the generated images. Interestingly, LidarCLIP with captions even outperforms image CLIP without captions, underscoring the effectiveness of our approach in generating high-quality images from point cloud data.



\begin{figure}[t]
\setlength{\tabcolsep}{1pt}
    \begin{tabular}{ccc}
        \multicolumn{3}{c}{Generated caption: A man walks on the street} \\
        \includegraphics[width=0.32 \linewidth]{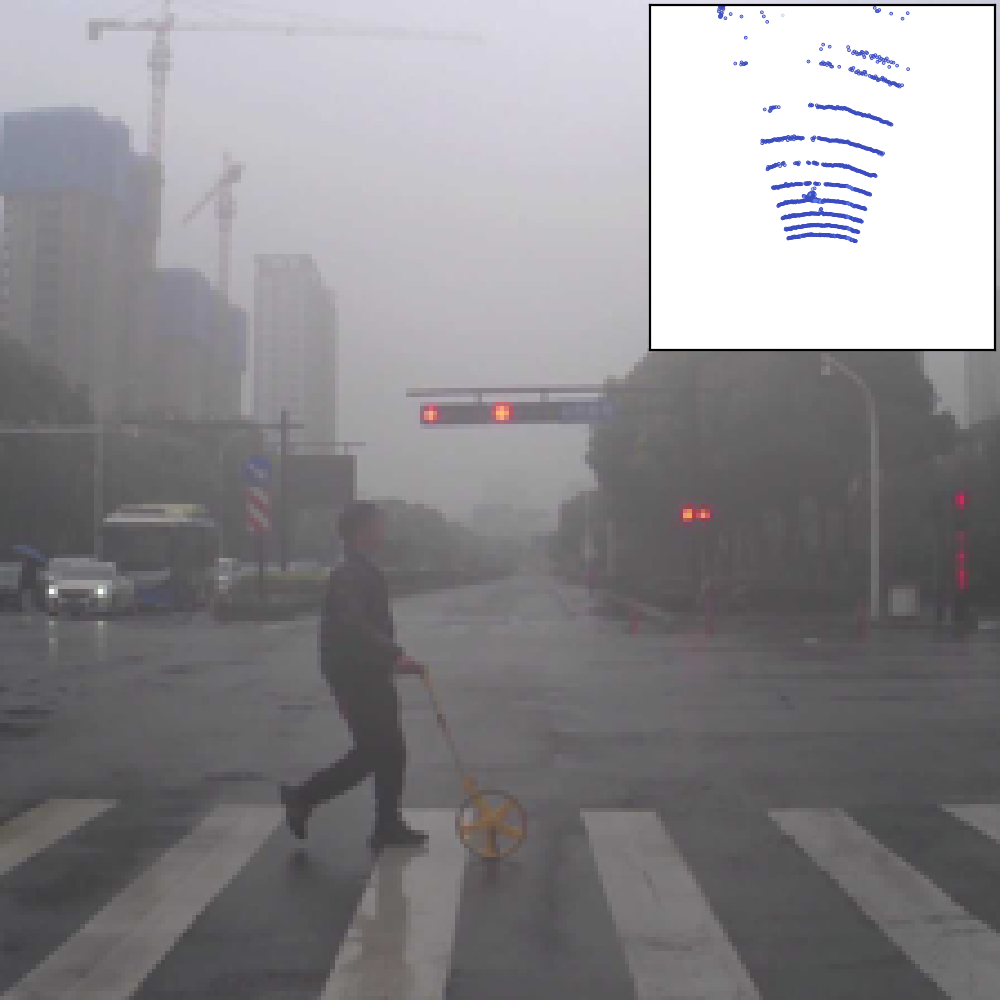} &
        \includegraphics[width=0.32 \linewidth, trim=0mm 0mm 5mm 0mm, clip]{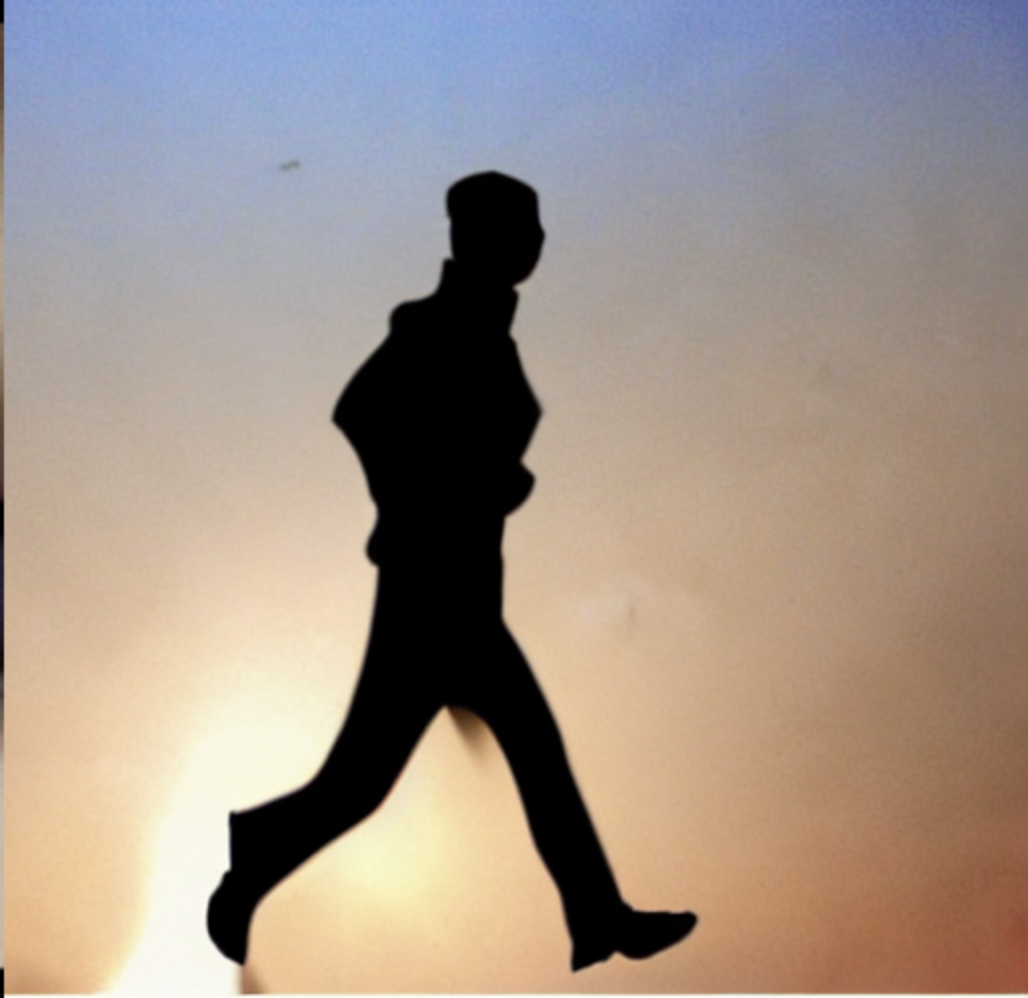} &
        \includegraphics[width=0.32 \linewidth]{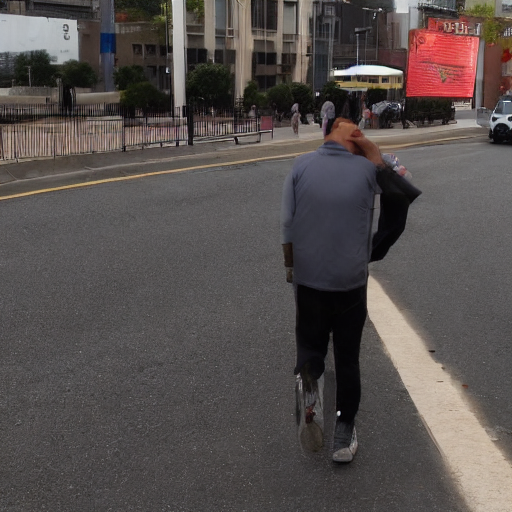} \\
        \multicolumn{3}{c}{Generated caption: A white car is driving down a street} \\
        \includegraphics[width=0.32 \linewidth]{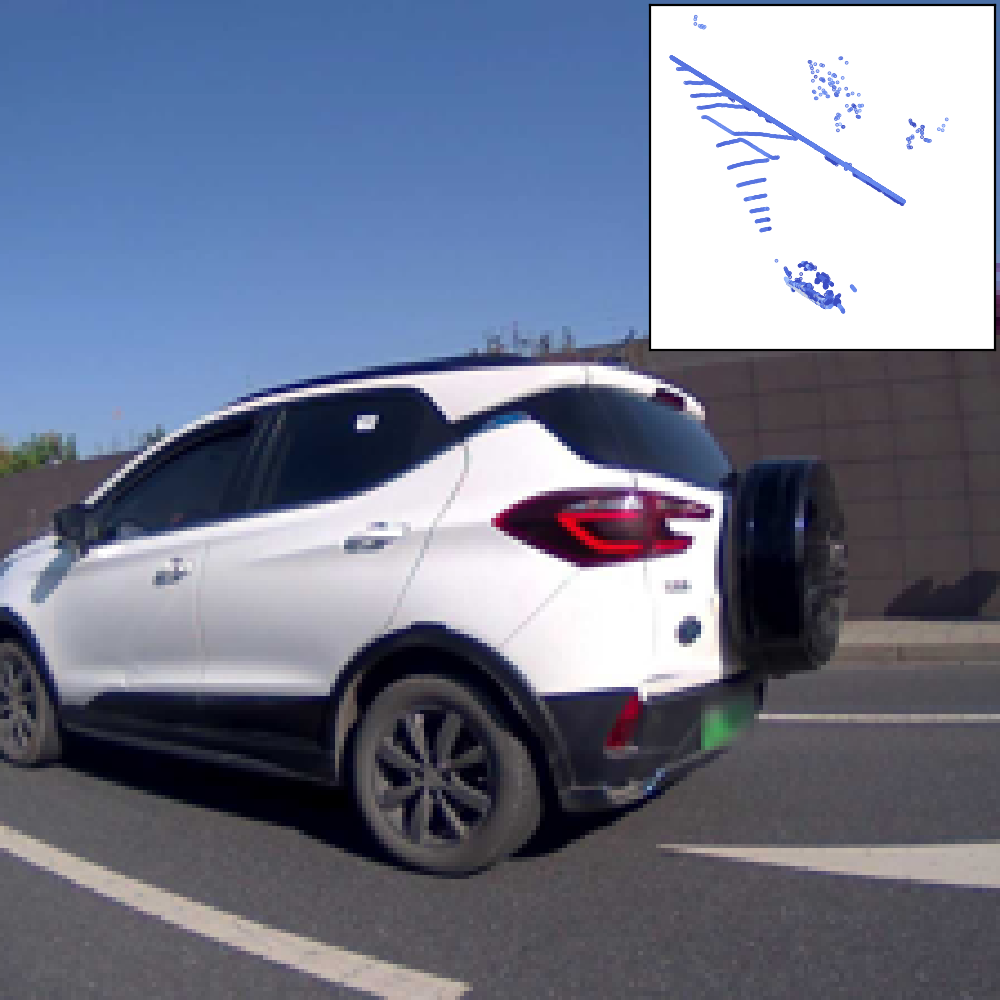} &
        \includegraphics[width=0.32 \linewidth]{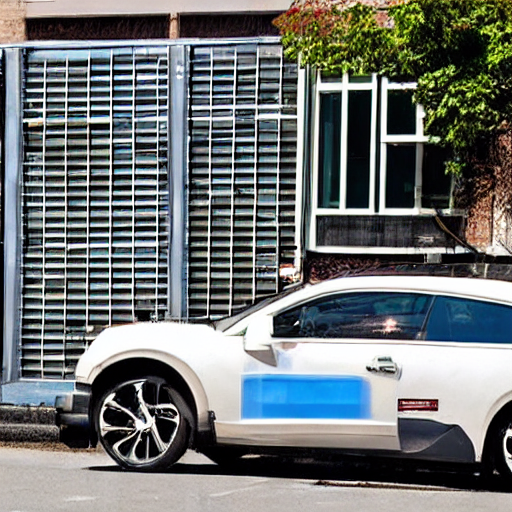} &
        \includegraphics[width=0.32 \linewidth]{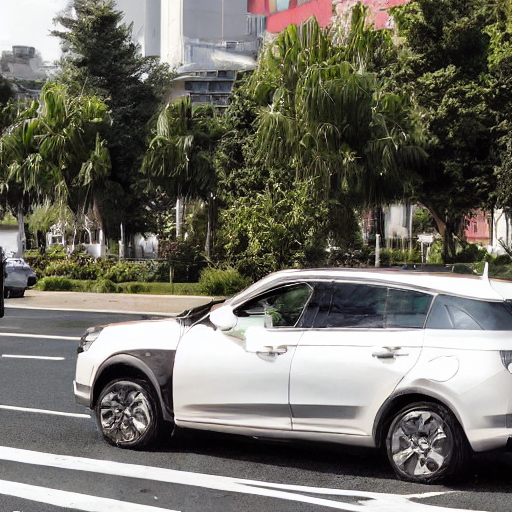} \\
    \end{tabular}
    \centering
    \caption{Example of generative application of LidarCLIP. A point cloud is embedded into the CLIP space (left, image only for reference) and used to generate text (top) and images. The image generation can be guided with only the lidar embedding (middle) or with both the lidar embedding and the generated caption (right).}
    \label{fig:generative_results}
\setlength{\tabcolsep}{6pt}
\end{figure}

Some qualitative results are shown in \cref{fig:generative_results}. We find that both generative tasks work fairly well out of the box. The generated images are not entirely realistic, partly due to a lack of tuning on our side, but there are clear similarities with the reference images. This demonstrates that our lidar embeddings can capture a surprising amount of detail. We hypothesize that guiding the diffusion process locally, by projecting regions of the point cloud into the image plane, would result in more realistic images. We hope that future work can investigate this avenue. Similarly, the captions can pick up the specifics of the scene. However, we notice that more `generic' images result in captions with very low diversity, such as ``several cars driving down a street next to tall buildings''. This is likely an artifact of the fact that the captioning model was trained on COCO, which only contains a few automotive images and has a limited vocabulary.

\begin{table}[]
    \centering
    \resizebox{0.99\linewidth}{!}{
        \begin{tabular}{l|lllllll}
            \toprule
                     & C (L) & C (I) & L    & I    & L+C  & I+C  & \cite{isola2017image} \\ \midrule
            FID $\downarrow$     & 83.0 & 81.7 & 68.7 & 58.7 & 53.7 & 46.9 & 114.2  \\
            CLIP-FID $\downarrow$ & 33.4 & 31.2 & 20.1 & 15.4 & 15.1 & 11.4 & 25.0  \\ \bottomrule
        \end{tabular}
    }
    \caption{FID and CLIP-FID (ViT-B/32) for $\approx$6k generated images from the ONCE \textit{val}. L=lidar, I=image, C (L/I)=caption only, from L/I.}
    \label{tab:image_generation}
\end{table}

\section{Limitations}
For the training of LidarCLIP, a single automotive dataset was used. While ONCE \cite{mao2021one} contains millions of image-lidar pairs, there are only around 1,000 densely sampled sequences, meaning that the dataset lacks diversity when compared to the 400 million text-image pairs used to train CLIP \cite{CLIP}. As an effect, LidarCLIP has mainly transferred CLIP's knowledge within an automotive setting and is not expected to work in a more general setting, such as indoors. Hence, an interesting future direction would be to train LidarCLIP on multiple datasets, with a variety of lidar sensors, scene conditions, and geographic locations. 

\section{Conclusions}
\label{sec:conclusions}
We propose LidarCLIP, which encodes lidar data into an existing text-image embedding space. Our method is trained using image-lidar pairs and enables multi-modal reasoning, connecting lidar to both text and images. While conceptually simple, LidarCLIP performs well over a range of tasks. For retrieval, we present a method for combining lidar and image features, outperforming their single-modality equivalents. Moreover, we use the joint retrieval method for finding challenging scenes under adverse sensor conditions. We also demonstrate that LidarCLIP enables several interesting applications off-the-shelf, including point cloud captioning and lidar-to-image generation. We hope LidarCLIP can inspire future work to dive deeper into connections between text and point cloud understanding, and explore tasks such as referring object detection and open-set semantic segmentation.

\section*{Acknowledgements}
This work was partially supported by the Wallenberg AI, Autonomous Systems and Software Program (WASP) funded by the Knut and Alice Wallenberg Foundation. Computational resources were provided by the Swedish National Infrastructure for Computing at \href{https://www.c3se.chalmers.se/}{C3SE} and \href{https://www.nsc.liu.se/}{NSC}, partially funded by the Swedish Research Council, grant agreement no. 2018-05973.

{\small
\bibliographystyle{ieee_fullname}
\bibliography{egbib}

\begin{thebibliography}{10}\itemsep=-1pt

\bibitem{aygun20214d}
Mehmet Aygun, Aljosa Osep, Mark Weber, Maxim Maximov, Cyrill Stachniss, Jens
  Behley, and Laura Leal-Taix{\'e}.
\newblock 4d panoptic lidar segmentation.
\newblock In {\em Proceedings of the IEEE/CVF Conference on Computer Vision and
  Pattern Recognition}, pages 5527--5537, 2021.

\bibitem{baldrati2022effective}
Alberto Baldrati, Marco Bertini, Tiberio Uricchio, and Alberto Del~Bimbo.
\newblock Effective conditioned and composed image retrieval combining
  clip-based features.
\newblock In {\em Proceedings of the IEEE/CVF Conference on Computer Vision and
  Pattern Recognition}, pages 21466--21474, 2022.

\bibitem{caesar2020nuscenes}
Holger Caesar, Varun Bankiti, Alex~H Lang, Sourabh Vora, Venice~Erin Liong,
  Qiang Xu, Anush Krishnan, Yu Pan, Giancarlo Baldan, and Oscar Beijbom.
\newblock nuscenes: A multimodal dataset for autonomous driving.
\newblock In {\em Proceedings of the IEEE/CVF conference on computer vision and
  pattern recognition}, pages 11621--11631, 2020.

\bibitem{carlssoncross}
Fredrik Carlsson, Philipp Eisen, Faton Rekathati, and Magnus Sahlgren.
\newblock Cross-lingual and multilingual clip.
\newblock In {\em Proceedings of the Thirteenth Language Resources and
  Evaluation Conference}, pages 6848--6854, 2022.

\bibitem{chen2020scanrefer}
Dave~Zhenyu Chen, Angel~X Chang, and Matthias Nie{\ss}ner.
\newblock Scanrefer: 3d object localization in rgb-d scans using natural
  language.
\newblock In {\em Computer Vision--ECCV 2020: 16th European Conference,
  Glasgow, UK, August 23--28, 2020, Proceedings, Part XX}, pages 202--221.
  Springer, 2020.

\bibitem{chen2020simple}
Ting Chen, Simon Kornblith, Mohammad Norouzi, and Geoffrey Hinton.
\newblock A simple framework for contrastive learning of visual
  representations.
\newblock In {\em International conference on machine learning}, pages
  1597--1607. PMLR, 2020.

\bibitem{cheraghian2022zero}
Ali Cheraghian, Shafin Rahman, Townim~F Chowdhury, Dylan Campbell, and Lars
  Petersson.
\newblock Zero-shot learning on 3d point cloud objects and beyond.
\newblock {\em International Journal of Computer Vision}, 130(10):2364--2384,
  2022.

\bibitem{dhariwal2021diffusion}
Prafulla Dhariwal and Alexander~Quinn Nichol.
\newblock Diffusion models beat {GAN}s on image synthesis.
\newblock In A. Beygelzimer, Y. Dauphin, P. Liang, and J.~Wortman Vaughan,
  editors, {\em Advances in Neural Information Processing Systems}, 2021.

\bibitem{9304812}
Mahdi Elhousni and Xinming Huang.
\newblock A survey on 3d lidar localization for autonomous vehicles.
\newblock In {\em 2020 IEEE Intelligent Vehicles Symposium (IV)}, pages
  1879--1884, 2020.

\bibitem{Fan_2022_CVPR}
Lue Fan, Ziqi Pang, Tianyuan Zhang, Yu-Xiong Wang, Hang Zhao, Feng Wang, Naiyan
  Wang, and Zhaoxiang Zhang.
\newblock Embracing single stride 3d object detector with sparse transformer.
\newblock In {\em Proceedings of the IEEE/CVF Conference on Computer Vision and
  Pattern Recognition}, pages 8458--8468, June 2022.

\bibitem{fan2022embracing}
Lue Fan, Ziqi Pang, Tianyuan Zhang, Yu-Xiong Wang, Hang Zhao, Feng Wang, Naiyan
  Wang, and Zhaoxiang Zhang.
\newblock Embracing single stride 3d object detector with sparse transformer.
\newblock In {\em Proceedings of the IEEE/CVF Conference on Computer Vision and
  Pattern Recognition}, pages 8458--8468, 2022.

\bibitem{Ganaie_2022}
M.A. Ganaie, Minghui Hu, A.K. Malik, M. Tanveer, and P.N. Suganthan.
\newblock Ensemble deep learning: A review.
\newblock {\em Engineering Applications of Artificial Intelligence},
  115:105151, oct 2022.

\bibitem{geng2020recent}
Chuanxing Geng, Sheng-jun Huang, and Songcan Chen.
\newblock Recent advances in open set recognition: A survey.
\newblock {\em IEEE transactions on pattern analysis and machine intelligence},
  43(10):3614--3631, 2020.

\bibitem{ghorab2013personalised}
M~Rami Ghorab, Dong Zhou, Alexander O’connor, and Vincent Wade.
\newblock Personalised information retrieval: survey and classification.
\newblock {\em User Modeling and User-Adapted Interaction}, 23(4):381--443,
  2013.

\bibitem{Gu2022OpenvocabularyOD}
Xiuye Gu, Tsung-Yi Lin, Weicheng Kuo, and Yin Cui.
\newblock Open-vocabulary object detection via vision and language knowledge
  distillation.
\newblock In {\em International Conference on Learning Representations}, 2022.

\bibitem{9747631}
Andrey Guzhov, Federico Raue, Jörn Hees, and Andreas Dengel.
\newblock Audioclip: Extending clip to image, text and audio.
\newblock In {\em IEEE International Conference on Acoustics, Speech and Signal
  Processing}, pages 976--980, 2022.

\bibitem{heinzler2019weather}
Robin Heinzler, Philipp Schindler, J{\"u}rgen Seekircher, Werner Ritter, and
  Wilhelm Stork.
\newblock Weather influence and classification with automotive lidar sensors.
\newblock In {\em 2019 IEEE intelligent vehicles symposium (IV)}, pages
  1527--1534. IEEE, 2019.

\bibitem{hendriksen2022extending}
Mariya Hendriksen, Maurits Bleeker, Svitlana Vakulenko, Nanne~van Noord, Ernst
  Kuiper, and Maarten~de Rijke.
\newblock Extending clip for category-to-image retrieval in e-commerce.
\newblock In {\em European Conference on Information Retrieval}, pages
  289--303. Springer, 2022.

\bibitem{huang2022clip2point}
Tianyu Huang, Bowen Dong, Yunhan Yang, Xiaoshui Huang, Rynson~WH Lau, Wanli
  Ouyang, and Wangmeng Zuo.
\newblock Clip2point: Transfer clip to point cloud classification with
  image-depth pre-training.
\newblock {\em arXiv preprint arXiv:2210.01055}, 2022.

\bibitem{isola2017image}
Phillip Isola, Jun-Yan Zhu, Tinghui Zhou, and Alexei~A Efros.
\newblock Image-to-image translation with conditional adversarial networks.
\newblock In {\em Proceedings of the IEEE conference on computer vision and
  pattern recognition}, pages 1125--1134, 2017.

\bibitem{jhaldiyal2022semantic}
Alok Jhaldiyal and Navendu Chaudhary.
\newblock Semantic segmentation of 3d lidar data using deep learning: a review
  of projection-based methods.
\newblock {\em Applied Intelligence}, pages 1--12, 2022.

\bibitem{adam}
Diederik~P. Kingma and Jimmy Ba.
\newblock Adam: {A} method for stochastic optimization.
\newblock In Yoshua Bengio and Yann LeCun, editors, {\em 3rd International
  Conference on Learning Representations, {ICLR} 2015, San Diego, CA, USA, May
  7-9, 2015, Conference Track Proceedings}, 2015.

\bibitem{kynkaanniemi2022role}
Tuomas Kynkäänniemi, Tero Karras, Miika Aittala, Timo Aila, and Jaakko
  Lehtinen.
\newblock The role of imagenet classes in fréchet inception distance.
\newblock In {\em Proceedings of International Conference on Learning
  Representations, {ICLR}}, 2023.

\bibitem{luddecke2022image}
Timo L{\"u}ddecke and Alexander Ecker.
\newblock Image segmentation using text and image prompts.
\newblock In {\em Proceedings of the IEEE/CVF Conference on Computer Vision and
  Pattern Recognition}, pages 7086--7096, 2022.

\bibitem{luo2022clip4clip}
Huaishao Luo, Lei Ji, Ming Zhong, Yang Chen, Wen Lei, Nan Duan, and Tianrui Li.
\newblock Clip4clip: An empirical study of clip for end to end video clip
  retrieval and captioning.
\newblock {\em Neurocomputing}, 508:293--304, 2022.

\bibitem{ma2022ei}
Haoyu Ma, Handong Zhao, Zhe Lin, Ajinkya Kale, Zhangyang Wang, Tong Yu,
  Jiuxiang Gu, Sunav Choudhary, and Xiaohui Xie.
\newblock Ei-clip: Entity-aware interventional contrastive learning for
  e-commerce cross-modal retrieval.
\newblock In {\em Proceedings of the IEEE/CVF Conference on Computer Vision and
  Pattern Recognition}, pages 18051--18061, 2022.

\bibitem{ma2022x}
Yiwei Ma, Guohai Xu, Xiaoshuai Sun, Ming Yan, Ji Zhang, and Rongrong Ji.
\newblock X-clip: End-to-end multi-grained contrastive learning for video-text
  retrieval.
\newblock In {\em Proceedings of the 30th ACM International Conference on
  Multimedia}, pages 638--647, 2022.

\bibitem{mao2021one}
Jiageng Mao, Minzhe Niu, Chenhan Jiang, hanxue liang, Jingheng Chen, Xiaodan
  Liang, Yamin Li, Chaoqiang Ye, Wei Zhang, Zhenguo Li, Jie Yu, Hang Xu, and
  Chunjing Xu.
\newblock One million scenes for autonomous driving: {ONCE} dataset.
\newblock In {\em Thirty-fifth Conference on Neural Information Processing
  Systems Datasets and Benchmarks Track (Round 1)}, 2021.

\bibitem{michele2021generative}
Bj{\"o}rn Michele, Alexandre Boulch, Gilles Puy, Maxime Bucher, and Renaud
  Marlet.
\newblock Generative zero-shot learning for semantic segmentation of 3d point
  clouds.
\newblock In {\em 2021 International Conference on 3D Vision (3DV)}, pages
  992--1002. IEEE, 2021.

\bibitem{Min2021JointPR}
Sewon Min, Kenton Lee, Ming-Wei Chang, Kristina Toutanova, and Hannaneh
  Hajishirzi.
\newblock Joint passage ranking for diverse multi-answer retrieval.
\newblock In {\em Proceedings of the 2021 Conference on Empirical Methods in
  Natural Language Processing}, pages 6997--7008, Online and Punta Cana,
  Dominican Republic, Nov. 2021. Association for Computational Linguistics.

\bibitem{clipcap}
Ron Mokady, Amir Hertz, and Amit~H Bermano.
\newblock Clipcap: Clip prefix for image captioning.
\newblock {\em arXiv preprint arXiv:2111.09734}, 2021.

\bibitem{parmar2022aliased_short}
Gaurav Parmar~et al.
\newblock On aliased resizing and surprising subtleties in {GAN} evaluation.
\newblock In {\em CVPR}, 2022.

\bibitem{CLIP}
Alec Radford, Jong~Wook Kim, Chris Hallacy, Aditya Ramesh, Gabriel Goh,
  Sandhini Agarwal, Girish Sastry, Amanda Askell, Pamela Mishkin, Jack Clark,
  et~al.
\newblock Learning transferable visual models from natural language
  supervision.
\newblock In {\em International Conference on Machine Learning}, pages
  8748--8763. PMLR, 2021.

\bibitem{dalle2}
Aditya Ramesh, Prafulla Dhariwal, Alex Nichol, Casey Chu, and Mark Chen.
\newblock Hierarchical text-conditional image generation with {CLIP} latents.
\newblock {\em arXiv preprint arXiv:2204.06125}, 2022.

\bibitem{rehman2012content}
Mehwish Rehman, Muhammad Iqbal, Muhammad Sharif, and Mudassar Raza.
\newblock Content based image retrieval: survey.
\newblock {\em World Applied Sciences Journal}, 19(3):404--412, 2012.

\bibitem{rombach2022high}
Robin Rombach, Andreas Blattmann, Dominik Lorenz, Patrick Esser, and Bj{\"o}rn
  Ommer.
\newblock High-resolution image synthesis with latent diffusion models.
\newblock In {\em Proceedings of the IEEE/CVF Conference on Computer Vision and
  Pattern Recognition}, pages 10684--10695, 2022.

\bibitem{rozenberszki2022language}
David Rozenberszki, Or Litany, and Angela Dai.
\newblock Language-grounded indoor 3d semantic segmentation in the wild.
\newblock In {\em Computer Vision--ECCV 2022: 17th European Conference, Tel
  Aviv, Israel, October 23--27, 2022, Proceedings, Part XXXIII}, pages
  125--141. Springer, 2022.

\bibitem{Sanghi_2022_CVPR}
Aditya Sanghi, Hang Chu, Joseph~G. Lambourne, Ye Wang, Chin-Yi Cheng, Marco
  Fumero, and Kamal~Rahimi Malekshan.
\newblock Clip-forge: Towards zero-shot text-to-shape generation.
\newblock In {\em Proceedings of the IEEE/CVF Conference on Computer Vision and
  Pattern Recognition}, pages 18603--18613, June 2022.

\bibitem{tang2021part2word}
Chuan Tang, Xi Yang, Bojian Wu, Zhizhong Han, and Yi Chang.
\newblock Part2word: Learning joint embedding of point clouds and text by
  matching parts to words.
\newblock {\em arXiv preprint arXiv:2107.01872}, 2021.

\bibitem{s20154306}
Jose~Roberto Vargas~Rivero, Thiemo Gerbich, Valentina Teiluf, Boris Buschardt,
  and Jia Chen.
\newblock Weather classification using an automotive lidar sensor based on
  detections on asphalt and atmosphere.
\newblock {\em Sensors}, 20(15), 2020.

\bibitem{Wang_2022_CVPR}
Can Wang, Menglei Chai, Mingming He, Dongdong Chen, and Jing Liao.
\newblock Clip-nerf: Text-and-image driven manipulation of neural radiance
  fields.
\newblock In {\em Proceedings of the IEEE/CVF Conference on Computer Vision and
  Pattern Recognition}, pages 3835--3844, June 2022.

\bibitem{wang2022cris}
Zhaoqing Wang, Yu Lu, Qiang Li, Xunqiang Tao, Yandong Guo, Mingming Gong, and
  Tongliang Liu.
\newblock Cris: Clip-driven referring image segmentation.
\newblock In {\em Proceedings of the IEEE/CVF Conference on Computer Vision and
  Pattern Recognition}, pages 11686--11695, 2022.

\bibitem{wu2022wav2clip}
Ho-Hsiang Wu, Prem Seetharaman, Kundan Kumar, and Juan~Pablo Bello.
\newblock Wav2clip: Learning robust audio representations from clip.
\newblock In {\em IEEE International Conference on Acoustics, Speech and Signal
  Processing}, pages 4563--4567. IEEE, 2022.

\bibitem{9181591}
Yutian Wu, Yueyu Wang, Shuwei Zhang, and Harutoshi Ogai.
\newblock Deep 3d object detection networks using lidar data: A review.
\newblock {\em IEEE Sensors Journal}, 21(2):1152--1171, 2021.

\bibitem{xu-etal-2021-videoclip}
Hu Xu, Gargi Ghosh, Po-Yao Huang, Dmytro Okhonko, Armen Aghajanyan, Florian
  Metze, Luke Zettlemoyer, and Christoph Feichtenhofer.
\newblock {VideoCLIP}: Contrastive pre-training for\\zero-shot video-text
  understanding.
\newblock In {\em Proceedings of the 2021 Conference on Empirical Methods in
  Natural Language Processing (EMNLP)}, Online, Nov. 2021. Association for
  Computational Linguistics.

\bibitem{zhang2022pointclip}
Renrui Zhang, Ziyu Guo, Wei Zhang, Kunchang Li, Xupeng Miao, Bin Cui, Yu Qiao,
  Peng Gao, and Hongsheng Li.
\newblock Pointclip: Point cloud understanding by {CLIP}.
\newblock In {\em Proceedings of the IEEE/CVF Conference on Computer Vision and
  Pattern Recognition}, pages 8552--8562, 2022.

\bibitem{zhao20213dvg}
Lichen Zhao, Daigang Cai, Lu Sheng, and Dong Xu.
\newblock 3dvg-transformer: Relation modeling for visual grounding on point
  clouds.
\newblock In {\em Proceedings of the IEEE/CVF International Conference on
  Computer Vision}, pages 2928--2937, 2021.

\bibitem{zhou2022extract}
Chong Zhou, Chen~Change Loy, and Bo Dai.
\newblock Extract free dense labels from clip.
\newblock In {\em Proceedings of the European Conference on Computer Vision},
  pages 696--712. Springer, 2022.

\end{thebibliography}
}

\clearpage
\appendix

\section{Supplementary material}

\section{Model details}
\cref{tab:sst_detatils} shows the hyperparameters for the SST encoder \cite{fan2022embracing} used for embedding point clouds in the CLIP space. Window shape refers to the number of voxels in each window. Other hyperparameters are used as is from the original implementation. 

The encoded voxel features from SST are further pooled with a multi-head self-attention layer to extract a single feature vector. Specifically, the CLIP embedding is initialized as the mean of all features and then attends to said features. The pooling uses 8 attention heads, learned positional embeddings, and a feature dimension that matches the current CLIP model, meaning 768 for ViT-L/14 and 512 for ViT-B/32. 

\begin{table}[b]
\centering
\begin{tabular}{@{}lll@{}}
\toprule
Parameter          & Value                   & Unit\\ \midrule
Voxel size         & (0.5, 0.5, 6)           & \si{\meter}\\
Window shape       & (12, 12, 1)             & -\\
Point cloud range  & (0, -20, -2, 40, 20, 4) & \si{\meter}\\
No. encoder layers & 4                       & -\\
$d$ model          & 128                     & -\\
$d$ feedforward    & 256                     & -\\ \bottomrule
\end{tabular}
\caption{Hyperparameters for SST encoder.}
\label{tab:sst_detatils}
\end{table}

\section{Training details}
All LidarCLIP models are trained on the union of the \textit{train} and \textit{raw\_large} splits which are defined in the ONCE development kit \cite{mao2021one}. The ViT-L/14 version is trained for 3 epochs, while ablations with ViT-B/32 used only 1 epoch for training. We use the Adam optimizer \cite{adam} with a base learning rate of $10^{-5}$. The learning rate follows a one-cycle learning rate scheduler with a maximum learning rate of $10^{-3}$ and cosine annealing and spends the first 10\% of the training increasing the learning rate. The training was done on four NVIDIA A100s with a batch size of 128, requiring about 27 hours for 3 epochs.

\section{Additional results}

\subsection{Retrieval}
In \cref{tab:retrieval-objects-nearby} we show class-wise performance when querying specifically for objects close to the ego-vehicle, as the main manuscript only displayed the average result. We find that lidar retrieval outperforms image retrieval for most classes, especially on Cyclists.

\begin{figure}[t]
    \centering
    \begin{tabular}{cc}
        \includegraphics[width=0.48\linewidth]{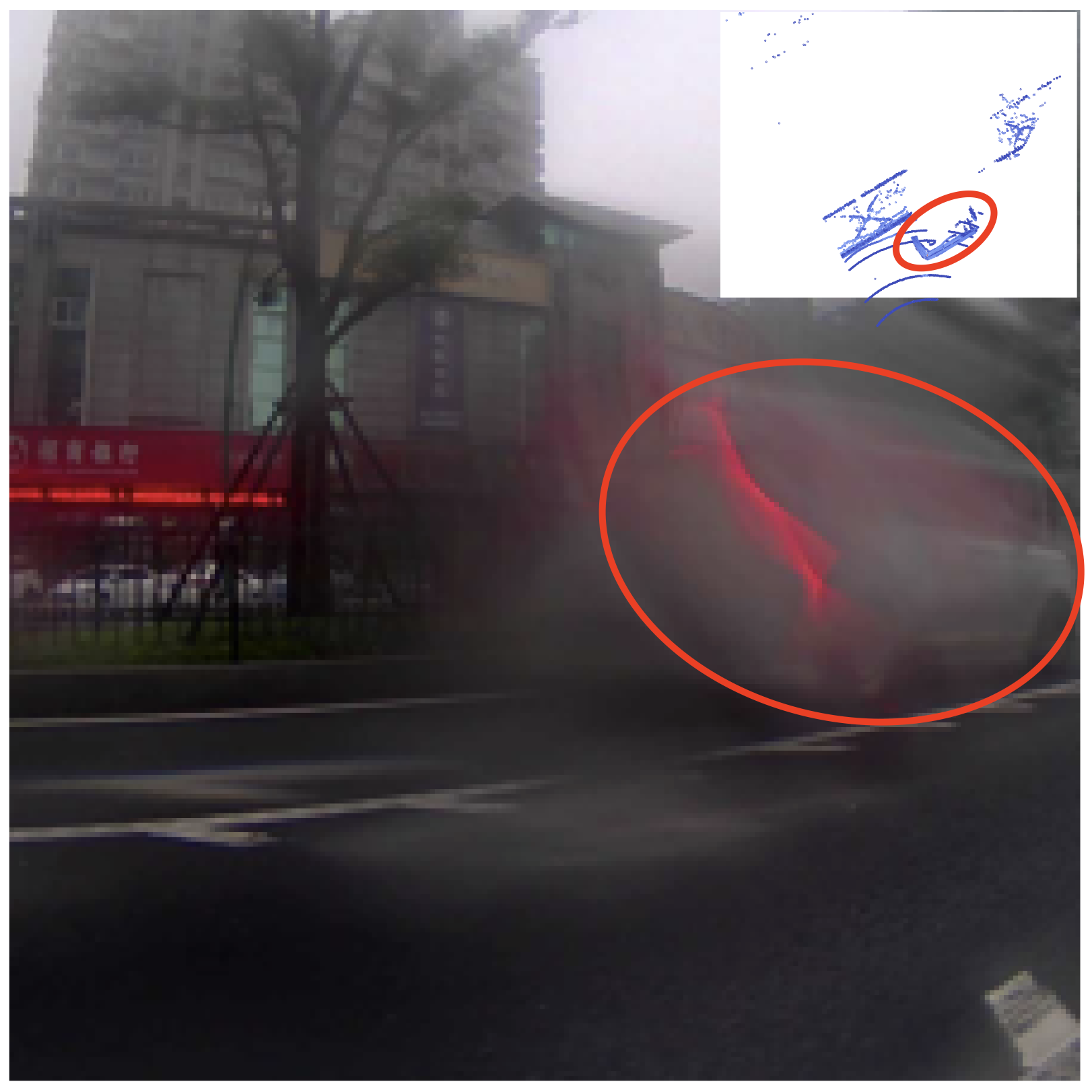} &
        \includegraphics[width=0.48\linewidth]{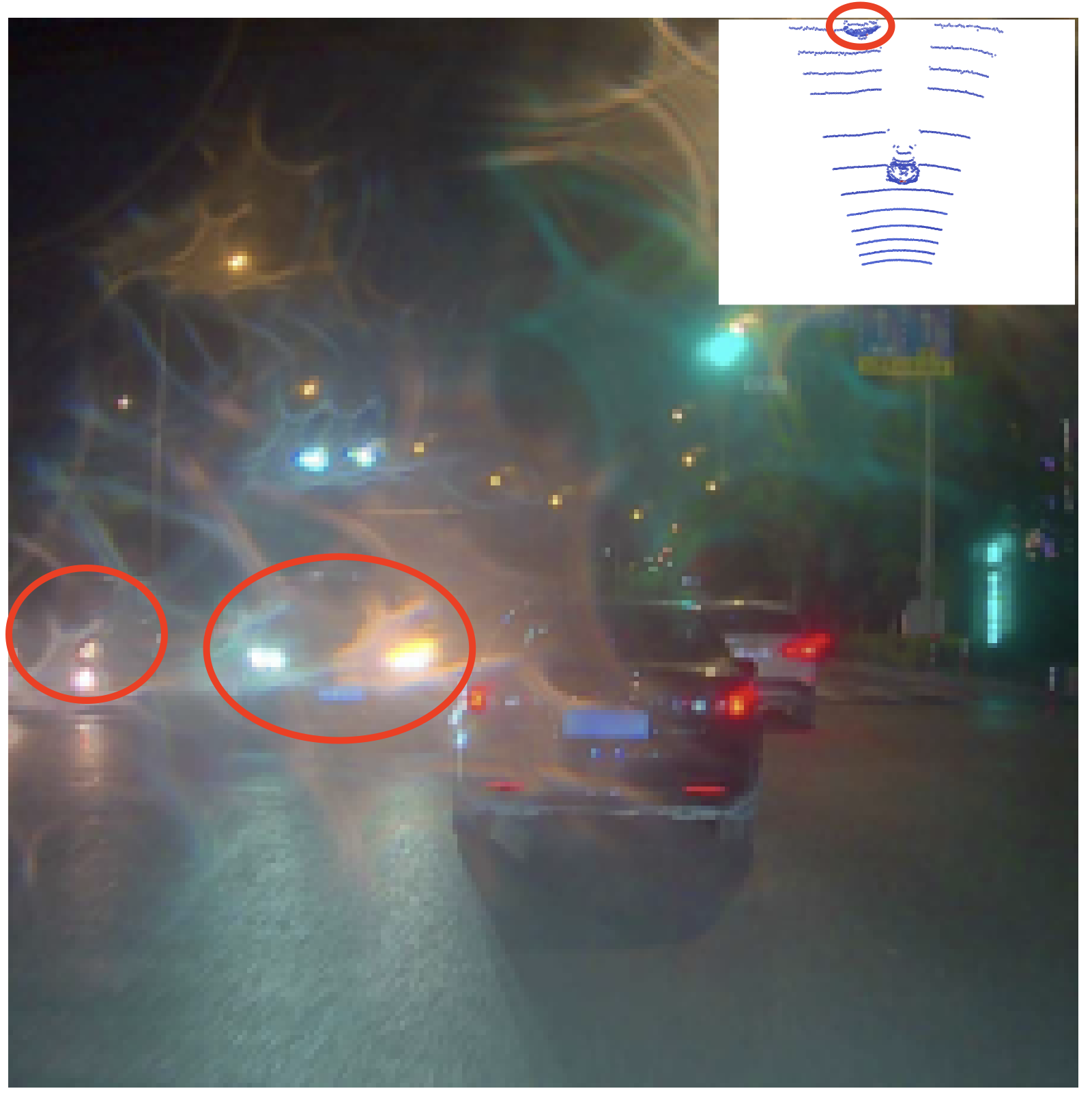}
    \end{tabular}
    \caption{Example of retrieval using separate prompts for image and lidar. We query for images with blur, water spray, glare, corruption, and lack of objects, and for point clouds with nearby trucks, pedestrians, cars, etc. By combining the scores of these separate queries, we can find edge cases that are extremely valuable during the training/validation of a camera-based perception system. These valuable objects are highlighted in red, both in the image and point cloud.}
    \label{fig:joint_blur_suppl}

\end{figure}

\noindent\textbf{Separate prompts.} \cref{fig:joint_blur_suppl} shows additional results for the joint retrieval with separate prompts for the image and lidar encoder. Again, these scenes are close to impossible to retrieve using a single modality.

\begin{figure*}[t]
    \begin{tabular}{ccccc}
        \multicolumn{5}{c}{``a bus''} \\
        \includegraphics[width=0.17 \linewidth]{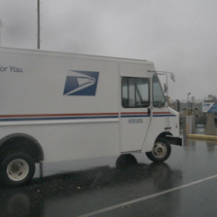} & \includegraphics[width=0.17 \linewidth]{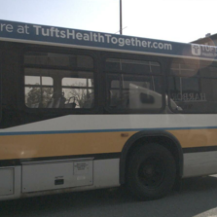} & \includegraphics[width=0.17 \linewidth]{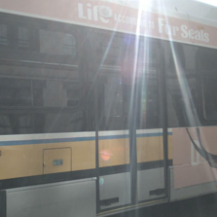} & \includegraphics[width=0.17 \linewidth]{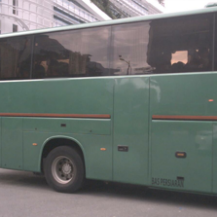} & \includegraphics[width=0.17 \linewidth]{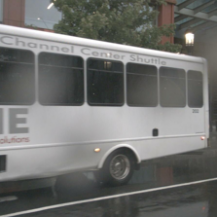} \\
        \multicolumn{5}{c}{``a nearby car''} \\
        \includegraphics[width=0.17 \linewidth]{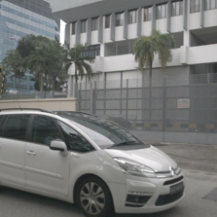} & \includegraphics[width=0.17 \linewidth]{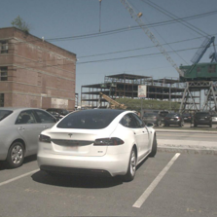} & \includegraphics[width=0.17 \linewidth]{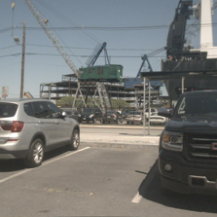} & \includegraphics[width=0.17 \linewidth]{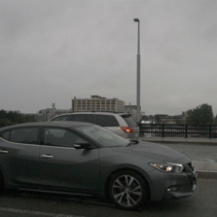} & \includegraphics[width=0.17 \linewidth]{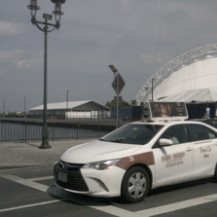} \\
        \multicolumn{5}{c}{``a distant car''} \\
        \includegraphics[width=0.17 \linewidth]{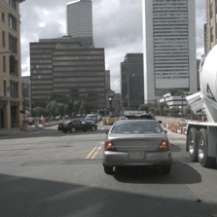} & \includegraphics[width=0.17 \linewidth]{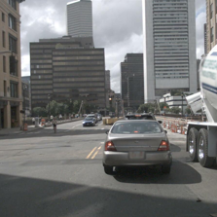} & \includegraphics[width=0.17 \linewidth]{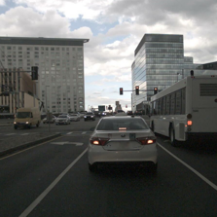} & \includegraphics[width=0.17 \linewidth]{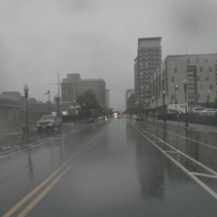} & \includegraphics[width=0.17 \linewidth]{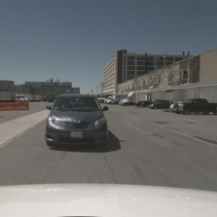} \\
        \multicolumn{5}{c}{``a person''} \\
        \includegraphics[width=0.17 \linewidth]{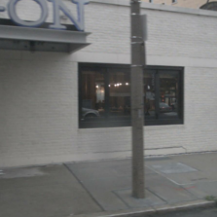} & \includegraphics[width=0.17 \linewidth]{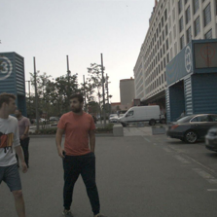} & \includegraphics[width=0.17 \linewidth]{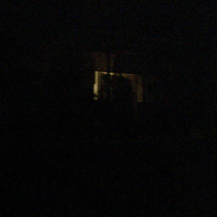} & \includegraphics[width=0.17 \linewidth]{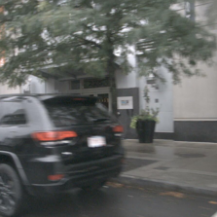} & \includegraphics[width=0.17 \linewidth]{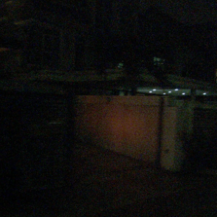} \\
        \multicolumn{5}{c}{``a person crossing the road''} \\
        \includegraphics[width=0.17 \linewidth]{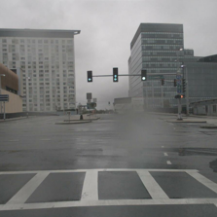} & \includegraphics[width=0.17 \linewidth]{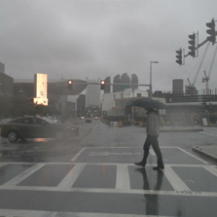} & \includegraphics[width=0.17 \linewidth]{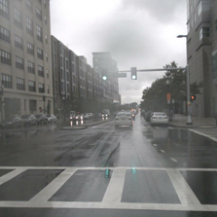} & \includegraphics[width=0.17 \linewidth]{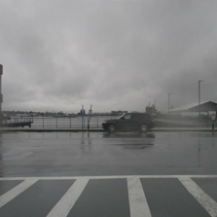} & \includegraphics[width=0.17 \linewidth]{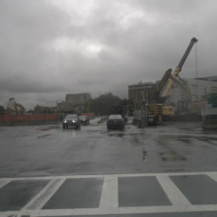} \\
        \multicolumn{5}{c}{``trees and bushes''} \\
        \includegraphics[width=0.17 \linewidth]{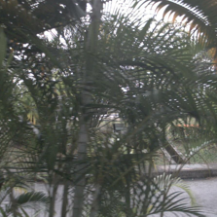} & \includegraphics[width=0.17 \linewidth]{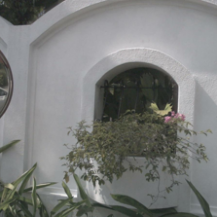} & \includegraphics[width=0.17 \linewidth]{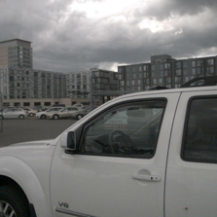} & \includegraphics[width=0.17 \linewidth]{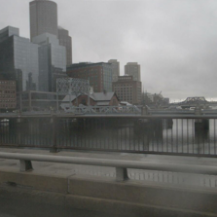} & \includegraphics[width=0.17 \linewidth]{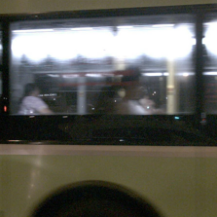} \\
    \end{tabular}
    \centering
    \caption{Top-5 retrieved examples when transferring LidarCLIP to nuScenes. LidarCLIP generalizes decently for distinct, large, objects, and even maintains some concept of distances. However, smaller objects and less uniform objects, like pedestrians and vegetation, do not transfer well.}
    \label{fig:nuscenes_suppl}
\end{figure*}

\noindent\textbf{nuScenes qualitative results.} In \cref{fig:nuscenes_suppl}, we show qualitative retrieval results on nuScenes using a ONCE-trained lidar encoder. As expected from the quantitative results, LidarCLIP does not generalize well overall. However, its performance is decent for distinct, large, objects, and it has some notion of distance.

\begin{table*}
    \centering
        \begin{tabular}{@{}lllllllllllll@{}}
        \toprule
        \multicolumn{1}{l|}{P@K}    & 10           & 100            & 10           & 100            & 10           & 100           & 10           & 100           & 10           & \multicolumn{1}{l|}{100}           &  10            & 100           \\ \midrule
        \multicolumn{1}{c}{}       & \multicolumn{2}{c}{Nearby Car}   & \multicolumn{2}{c}{Nearby Truck} & \multicolumn{2}{c}{Nearby Bus}   & \multicolumn{2}{c}{Nearby Ped.}  & \multicolumn{2}{c}{Nearby Cyclist}                    & \multicolumn{2}{|c}{Avg. Nearby}              \\ \midrule
        \multicolumn{1}{l|}{Image}  & \textbf{1.0} & 0.95           & 0.7          & 0.83           & 0.6          & 0.55            & \textbf{1.0} & 0.66           & 0.5          & \multicolumn{1}{l|}{0.37}                     & 0.76          & 0.67          \\
        \multicolumn{1}{l|}{Lidar}  & \textbf{1.0} & \textbf{0.98} & \textbf{0.8} & 0.70          & \textbf{1.0} & \textbf{0.90}  & 0.7          & 0.71          & \textbf{0.9} & \multicolumn{1}{l|}{\textbf{0.61}} & 0.88          & 0.78          \\
        \multicolumn{1}{l|}{Joint} & 0.9          & \textbf{0.98} & \textbf{0.8} & \textbf{0.92} & \textbf{1.0} & 0.79          & \textbf{1.0} & \textbf{0.84} & 0.8          & \multicolumn{1}{l|}{0.53}            & \textbf{0.90}  & \textbf{0.81} \\ \bottomrule
        \end{tabular}
    \caption{Retrieval for various object categories. We report precision at ranks 1, 10, and 100.}
    \label{tab:retrieval-objects-nearby}
\end{table*}

\subsection{Zero-shot scene classification}

Here, we provide additional examples of zero-shot classification using LidarCLIP. However, rather than object-level classification, we do zero-shot classification on entire scenes. We compare the performance of image-only, lidar-only, and the joint approach. 

\begin{figure*}[b]
    \centering
    \begin{tabular}{cc}
        \includegraphics[width=0.35 \linewidth]{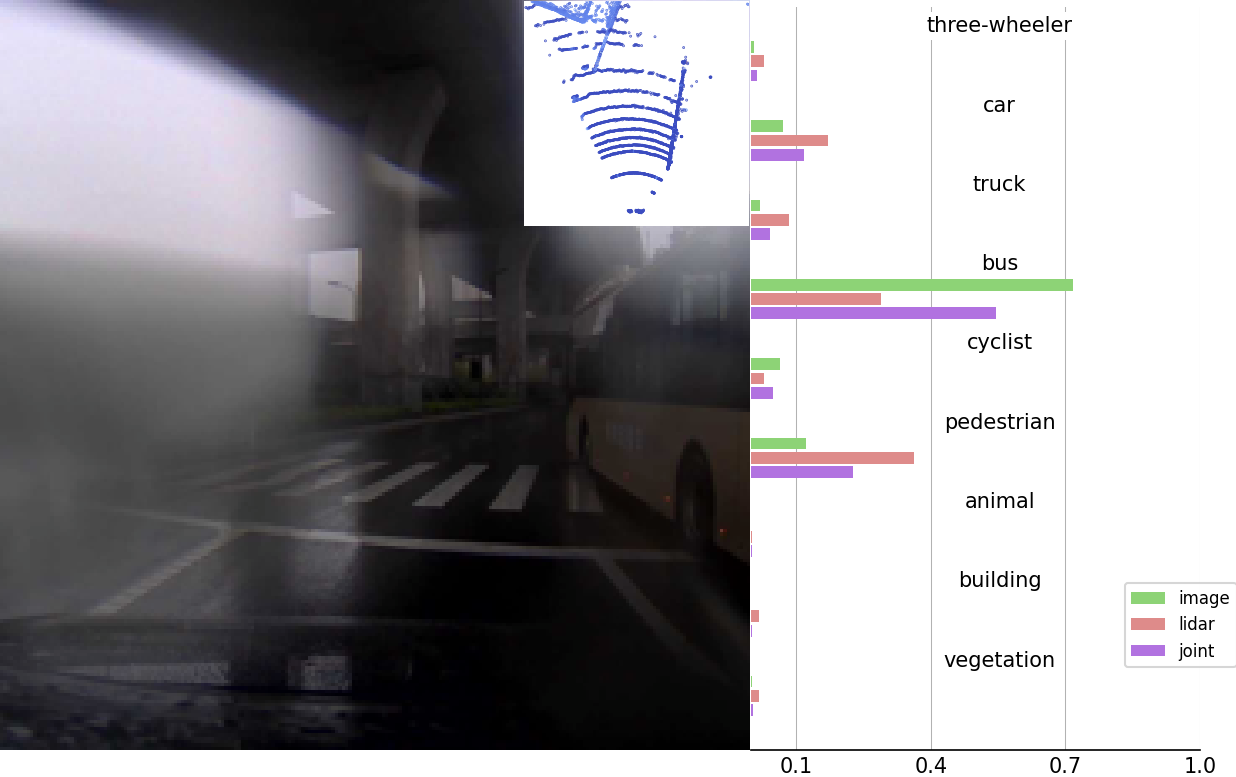} & \includegraphics[width=0.35 \linewidth]{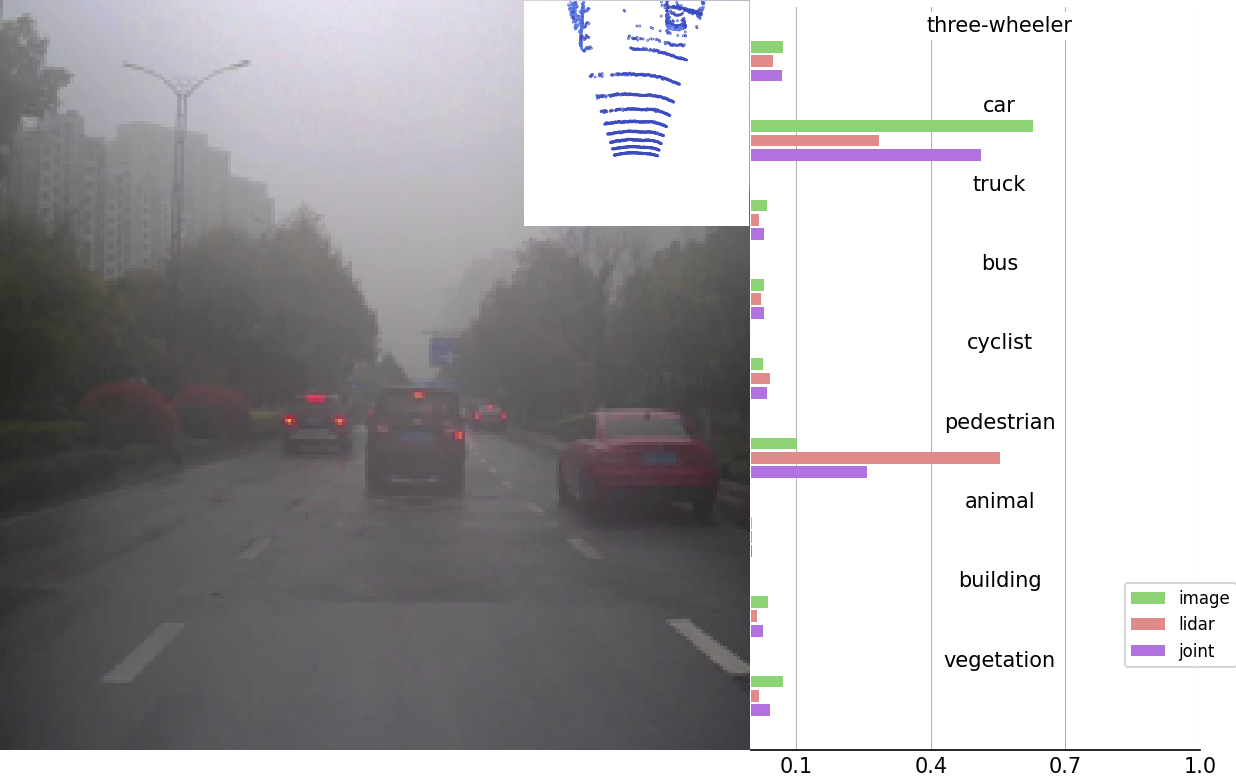} \\
        \includegraphics[width=0.35 \linewidth]{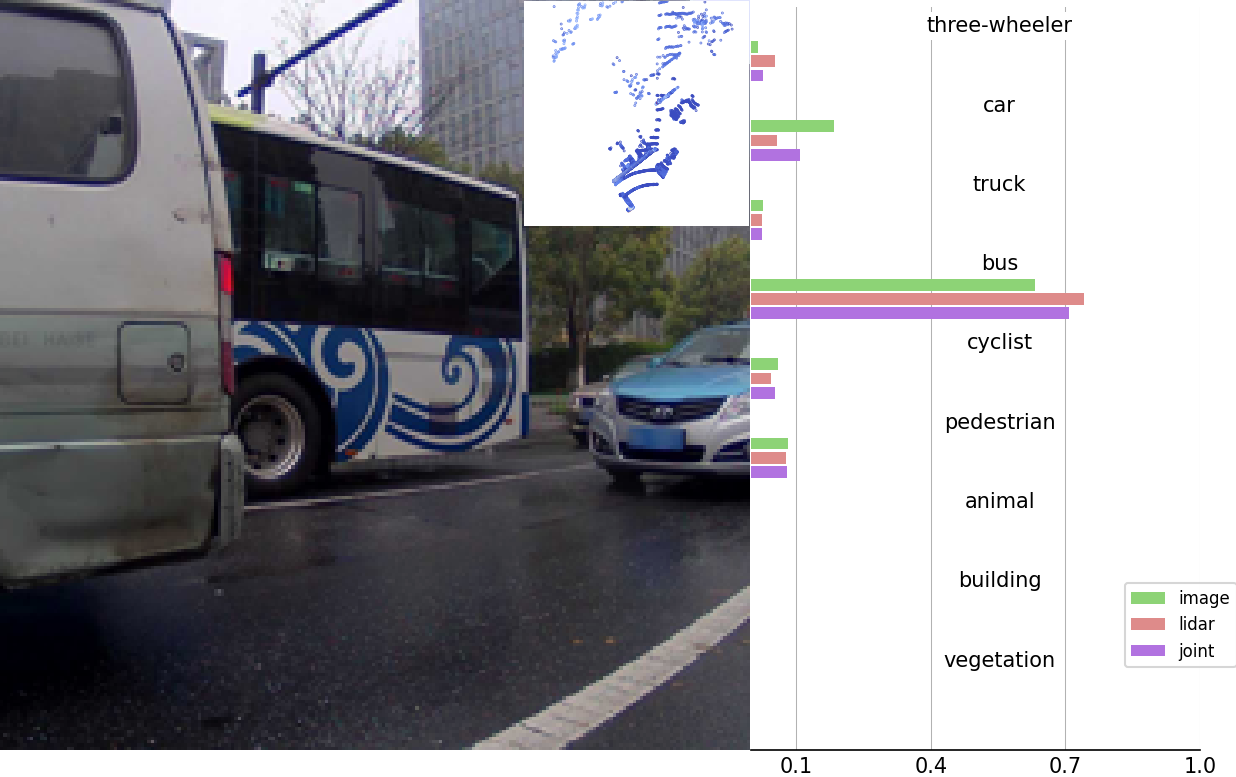} & \includegraphics[width=0.35 \linewidth]{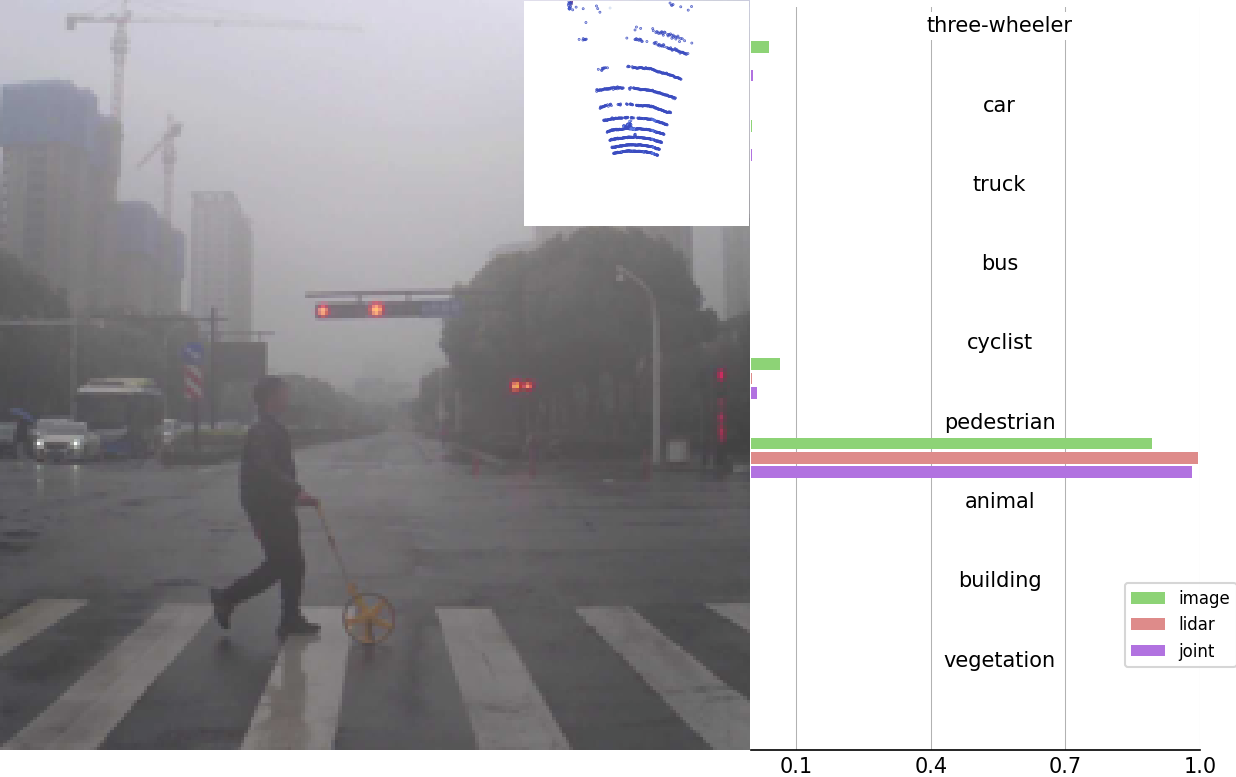} \\
        \includegraphics[width=0.35 \linewidth]{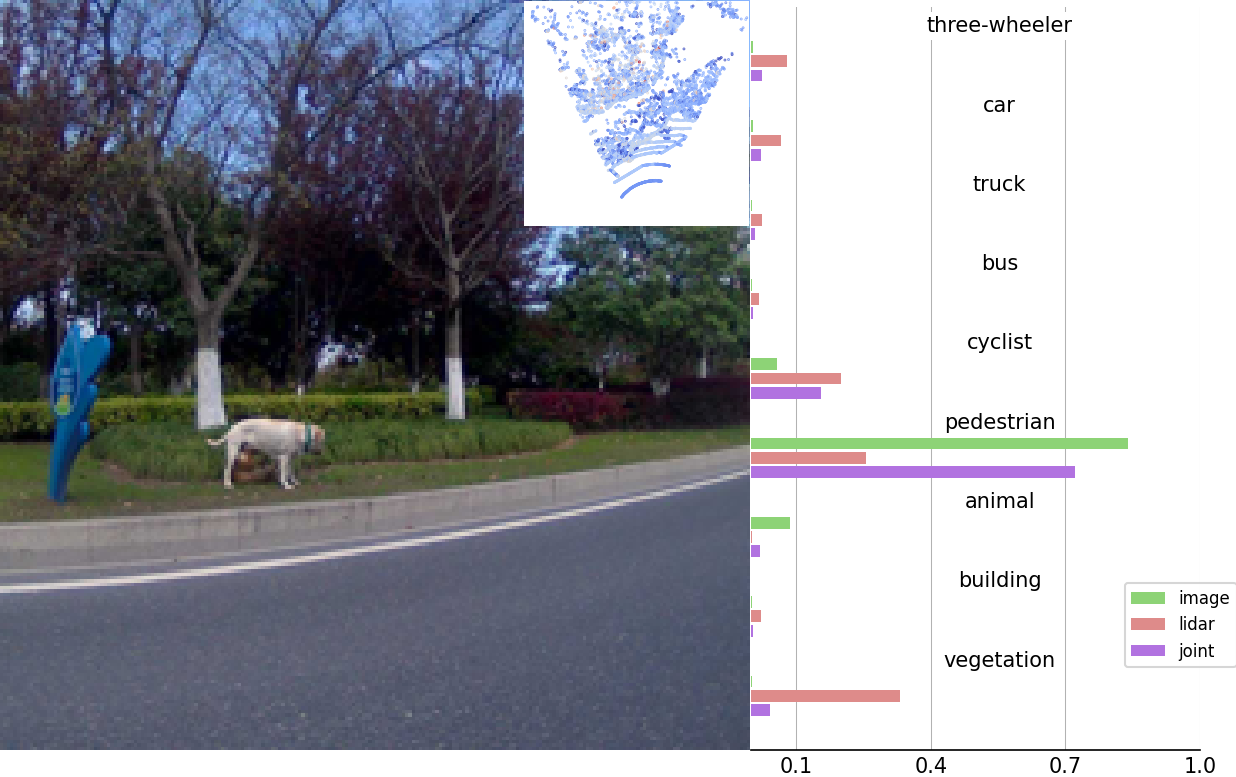} & \includegraphics[width=0.35 \linewidth]{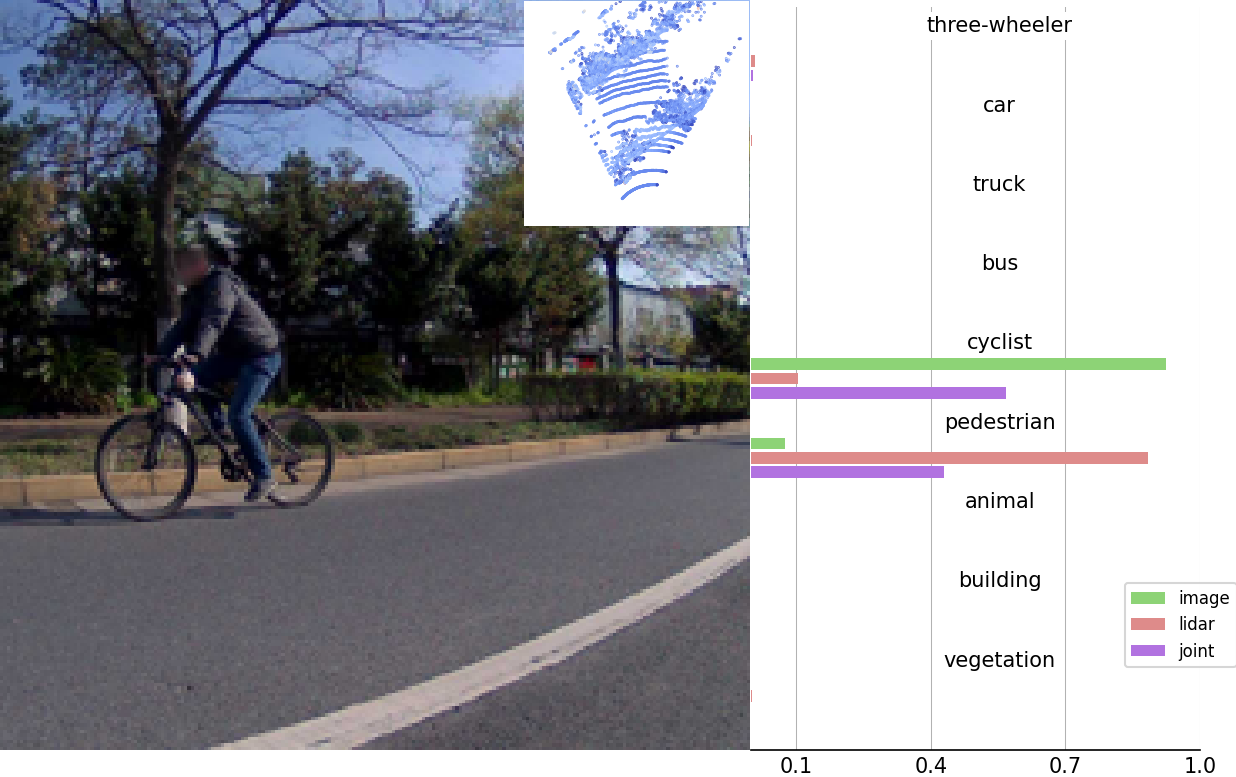} \\
        \includegraphics[width=0.35 \linewidth]{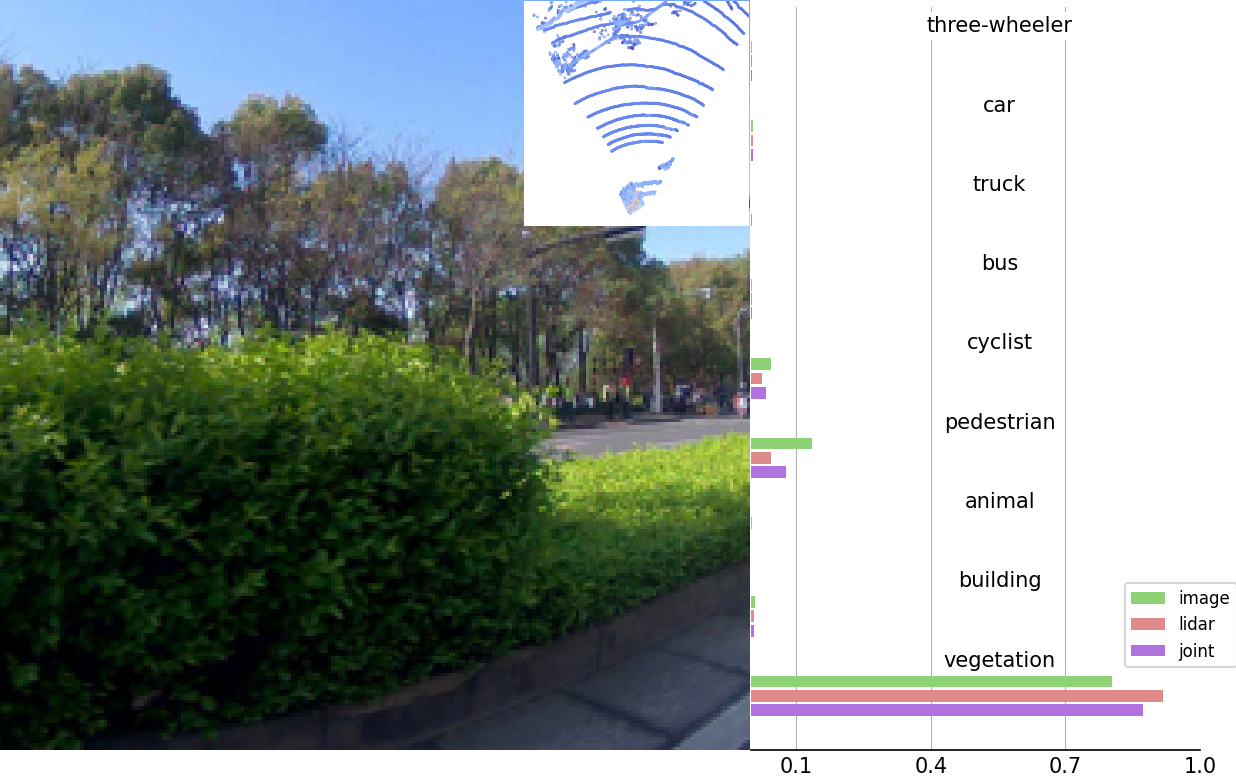} & \includegraphics[width=0.35 \linewidth]{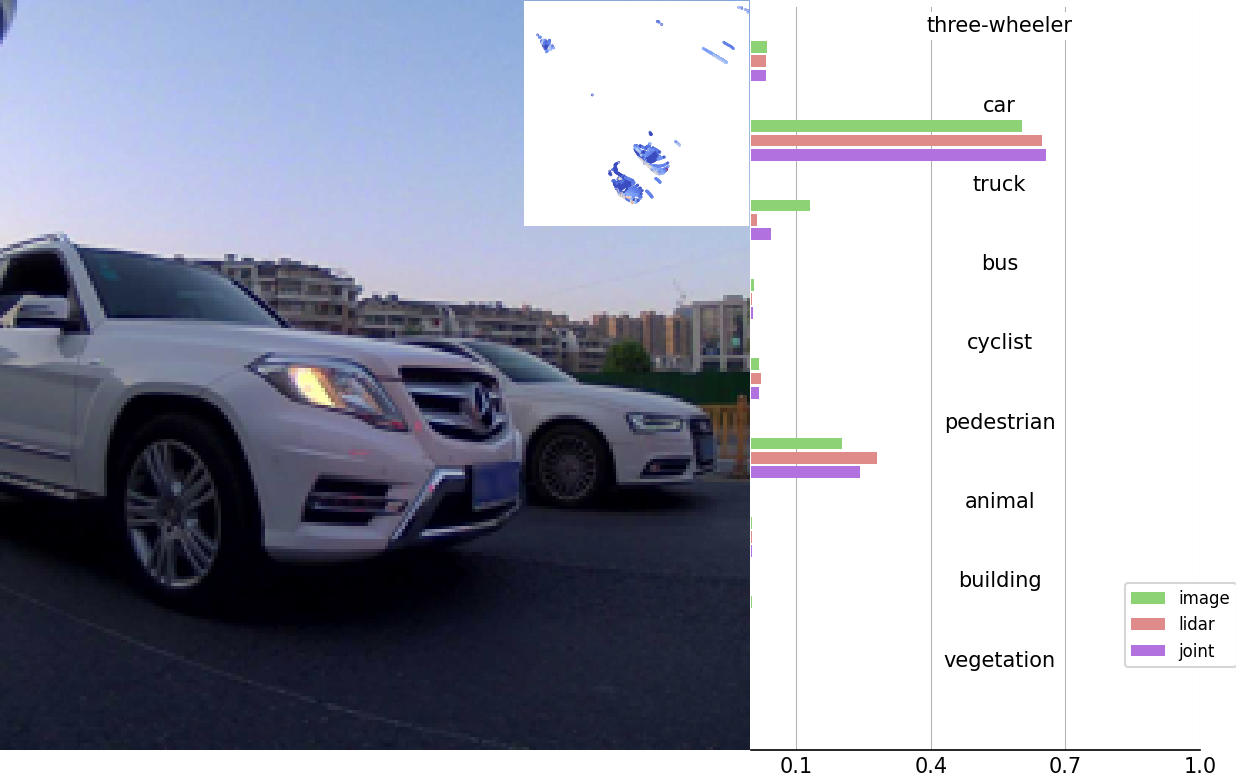} \\
        \includegraphics[width=0.35 \linewidth]{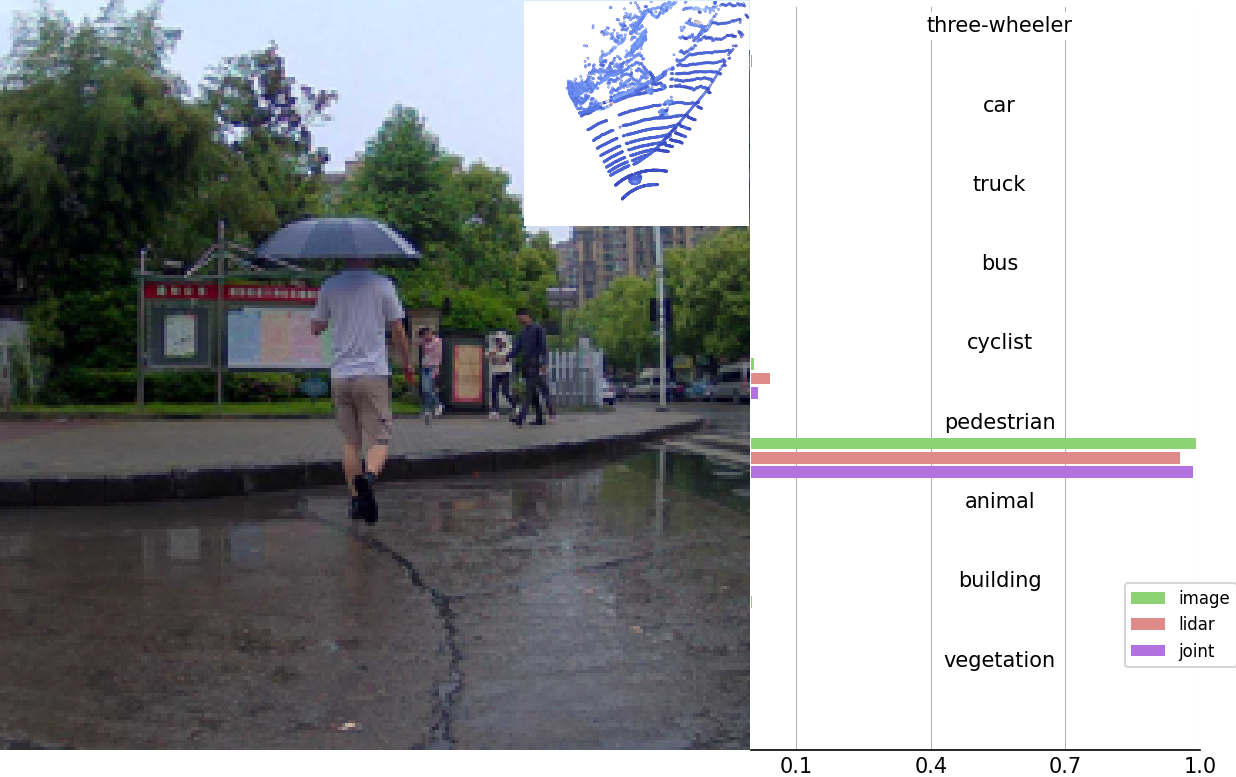} & \includegraphics[width=0.35 \linewidth]{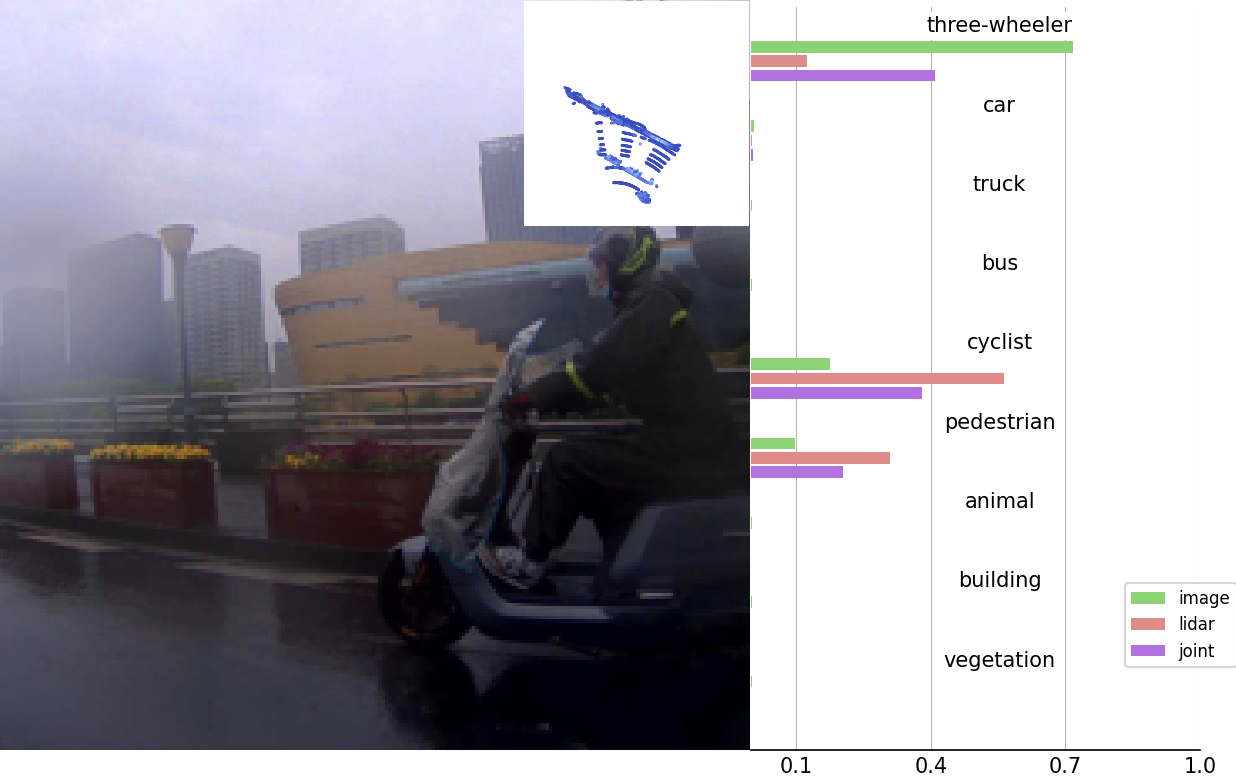} \\
    \end{tabular}
    \caption{Qualitative zero-shot classification on the ONCE validation/test set.}
    \label{fig:zero_shot_class_suppl}
\end{figure*}

\cref{fig:zero_shot_class_suppl} shows a diverse set of samples from the validation and test set. In many cases, LidarCLIP and image-based CLIP give similar results, highlighting the transfer of knowledge to the lidar domain. In some cases, however, the two models give contradictory classifications. For instance, LidarCLIP misclassifies the cyclist as a pedestrian, potentially due to the upright position and fewer points for the bike than the person. While their disagreement influences the joint method, ``cyclist'' remains the dominating class. Another interesting example is the image of the dog. None of the models manage to confidently classify the presence of an animal in the scene. This also highlights a shortcoming with our approach, where the image encoder's capacity may limit what the lidar encoder can learn. This can be circumvented to some extent by using even larger datasets, but a more effective approach could be local supervision. For instance, using CLIP features on a patch level to supervise frustums of voxel features, thus improving the understanding of fine-grained details. 

\begin{figure*}[tb]
    \centering
    \begin{tabular}{cc}
        \includegraphics[width=0.4 \linewidth]{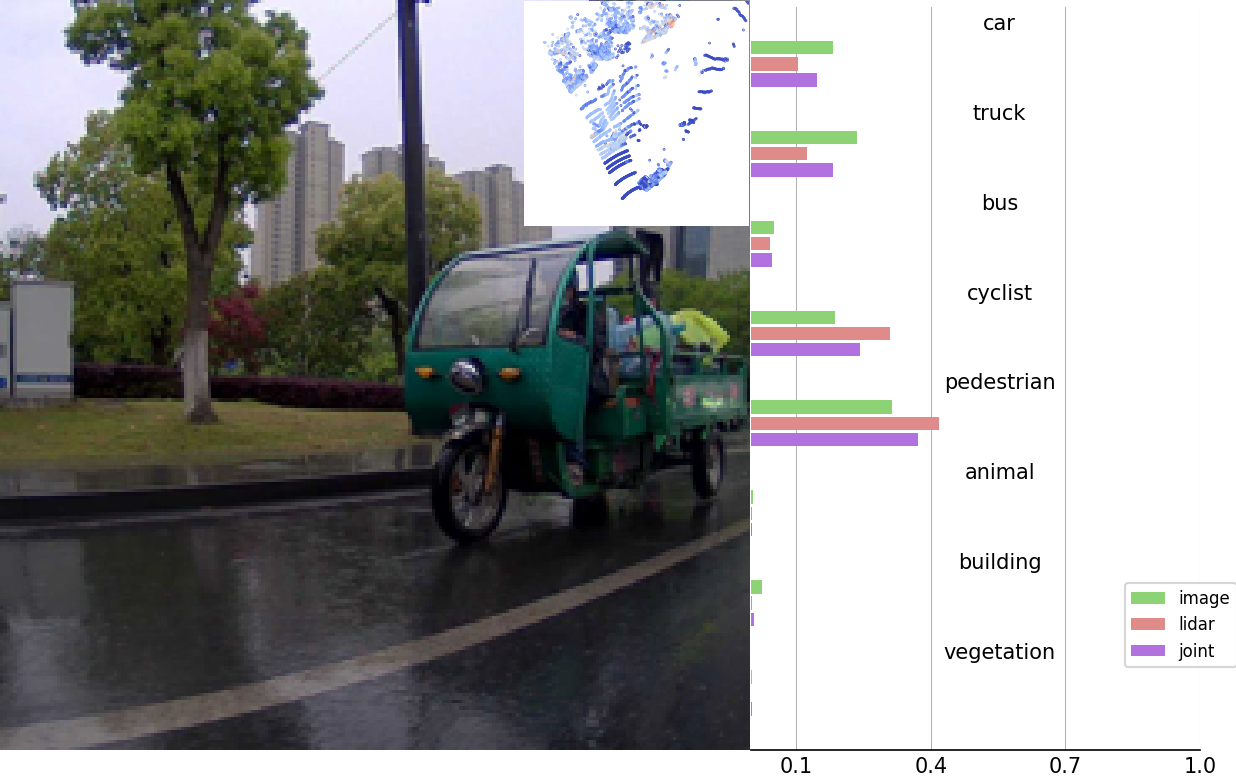} & \includegraphics[width=0.4 \linewidth]{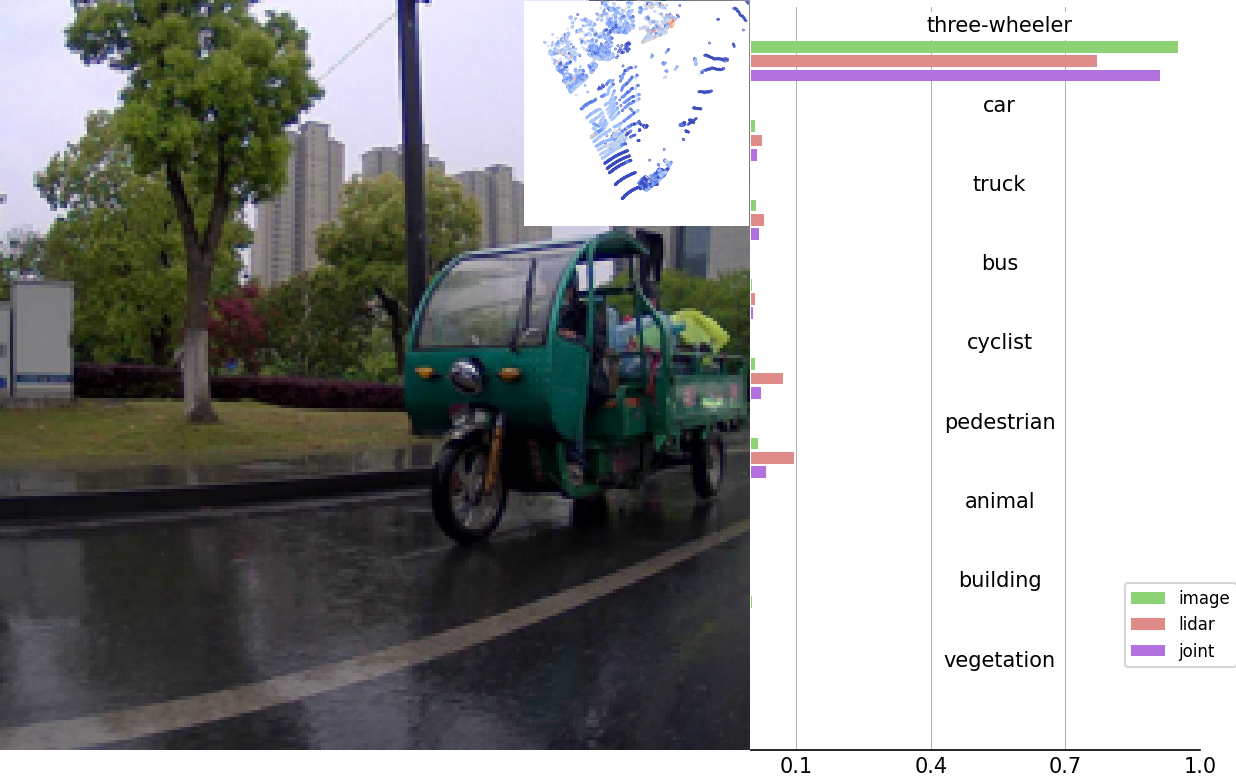}
    \end{tabular}
    \caption{Example of zero-shot classification that demonstrates the importance of picking good class prompts. When excluding the most appropriate class, ``three-wheeler'', the model is highly confused between the remaining classes.}
    \label{fig:zero_shot_missing_class_suppl}
\end{figure*}

In \cref{fig:zero_shot_missing_class_suppl} we highlight the importance, and problem, of including reasonable classes for zero-shot classification. The example scene contains a three-wheeler driving down the street. For the left sub-figure, none of the text prompts contains the word three-wheeler. Consequently, the model is confused between car, truck, cyclist, and pedestrian as none of these are perfect for describing the scene. When including ``three-wheeler'' as a separate class in the right sub-figure, the model accurately classifies the main subject of the image. To avoid such issues, we would like to create a class for unclassified or unknown objects, such that the model can express that none of the provided prompts is a good fit. Optimally, the model should be able to express what this class is, either by providing a caption or retrieving similar scenes, which can guide a human in naming and including additional classes. We hope that future work, potentially inspired by open-set recognition \cite{geng2020recent}, can study this more closely.  

\setlength{\tabcolsep}{6pt}
\setlength{\tabcolsep}{6pt}

\setlength{\tabcolsep}{1pt}
\setlength{\tabcolsep}{6pt}

\subsection{LidarCLIP for lidar sensing capabilities}

\begin{figure*}[t]
\centering
    \begin{tabular}{ccccc}
        \multicolumn{5}{c}{``a blue car''} \\
        \includegraphics[width=0.19 \linewidth]{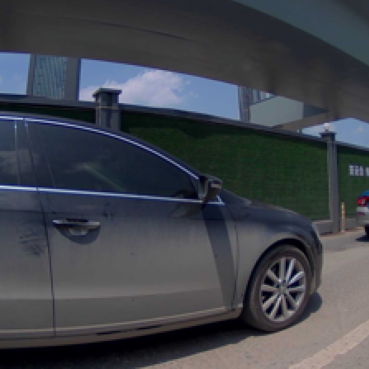} & \includegraphics[width=0.19 \linewidth]{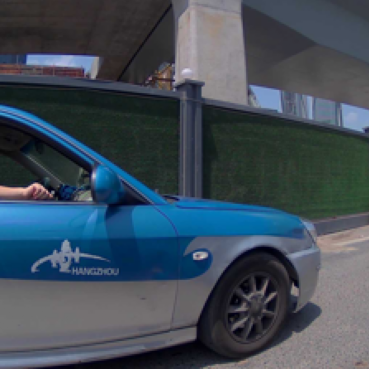} & \includegraphics[width=0.19 \linewidth]{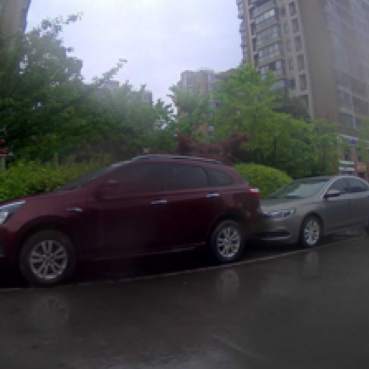} & \includegraphics[width=0.19 \linewidth]{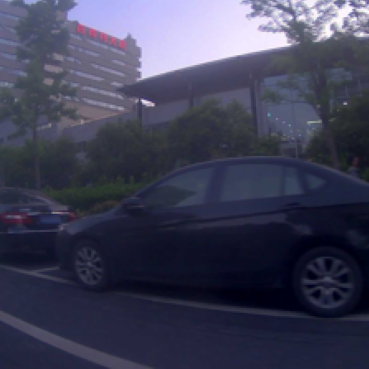} & \includegraphics[width=0.19 \linewidth]{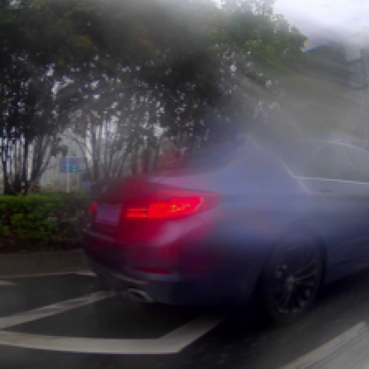} \\
        \multicolumn{5}{c}{``a green car''} \\
        \includegraphics[width=0.19 \linewidth]{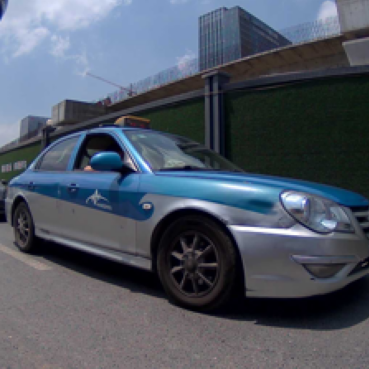} & \includegraphics[width=0.19 \linewidth]{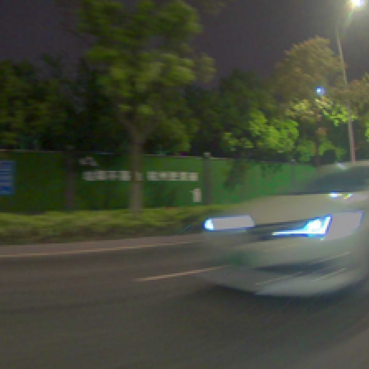} & \includegraphics[width=0.19 \linewidth]{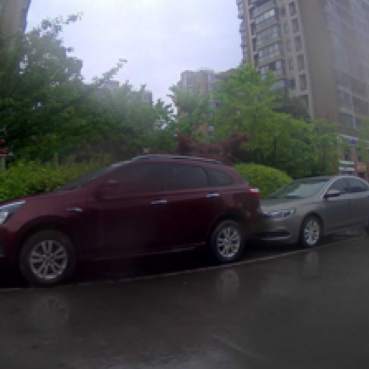} & \includegraphics[width=0.19 \linewidth]{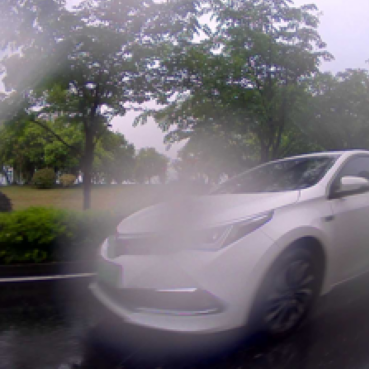} & \includegraphics[width=0.19 \linewidth]{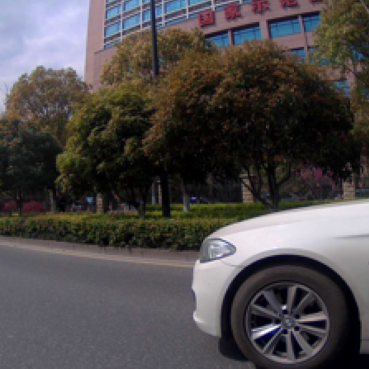} \\
        \multicolumn{5}{c}{``a gray car''} \\
        \includegraphics[width=0.19 \linewidth]{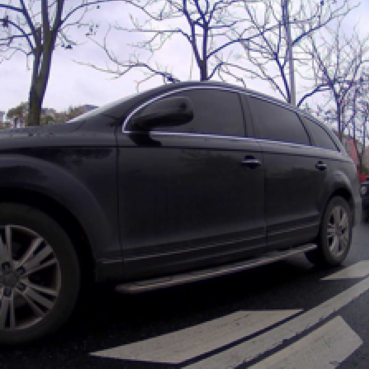} & \includegraphics[width=0.19 \linewidth]{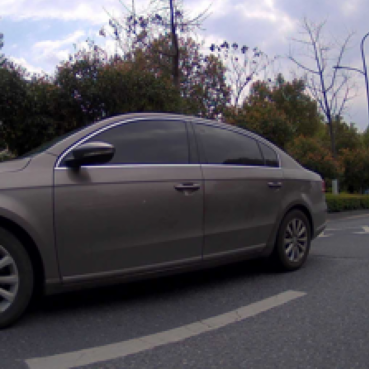} & \includegraphics[width=0.19 \linewidth]{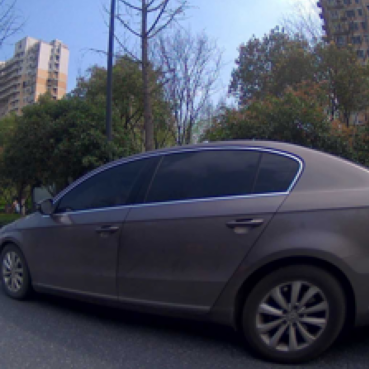} & \includegraphics[width=0.19 \linewidth]{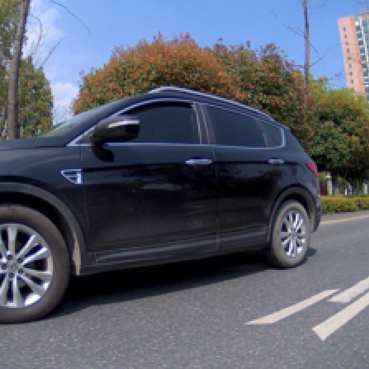} & \includegraphics[width=0.19 \linewidth]{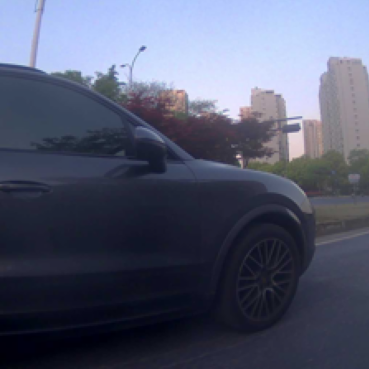} \\
        \multicolumn{5}{c}{``a yellow car''} \\
        \includegraphics[width=0.19 \linewidth]{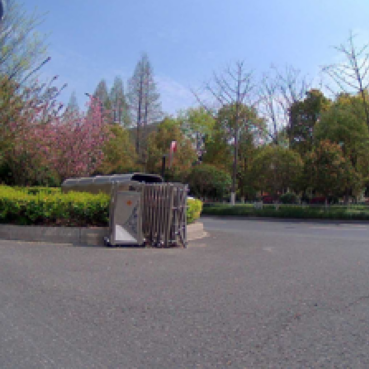} & \includegraphics[width=0.19 \linewidth]{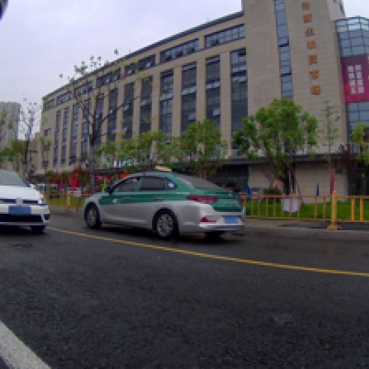} & \includegraphics[width=0.19 \linewidth]{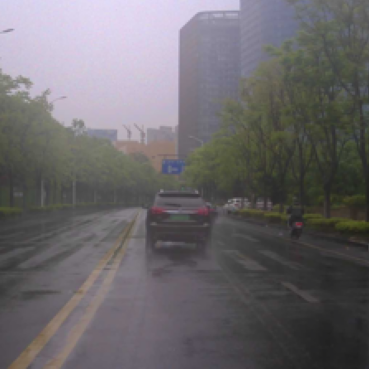} & \includegraphics[width=0.19 \linewidth]{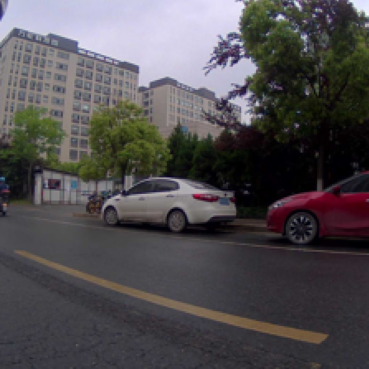} & \includegraphics[width=0.19 \linewidth]{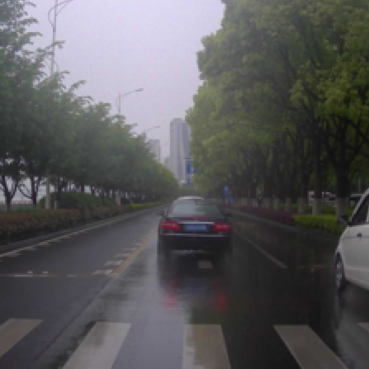}
    \end{tabular}
    \centering
    \caption{Top-5 retrieved examples from LidarCLIP for different colors. Note that we show images only for visualization, point clouds were used for retrieval. }
    \label{fig:color_retrieval_suppl}
\end{figure*}

In \cref{fig:color_retrieval_suppl}, we show additional examples of retrieved scenes for various colors. Note that images are only shown for reference. Similar to results in Sec. 4.2, LidarCLIP has no understanding of distinct colors but can discriminate between dark and bright. For instance, ``a gray car'' returns dark grey cars, while none of the cars for ``a yellow car'' are yellow. 

\setlength{\tabcolsep}{1pt}
\setlength{\tabcolsep}{6pt}

\setlength{\tabcolsep}{1pt}
\setlength{\tabcolsep}{6pt}

\subsection{Lidar to image and text}

\begin{figure*}[t]
\setlength{\tabcolsep}{1pt}
    \begin{tabular}{cc}
        \multicolumn{2}{c}{Generated caption: A city street with traffic lights and street signs.} \\
        \includegraphics[width=0.19 \linewidth]{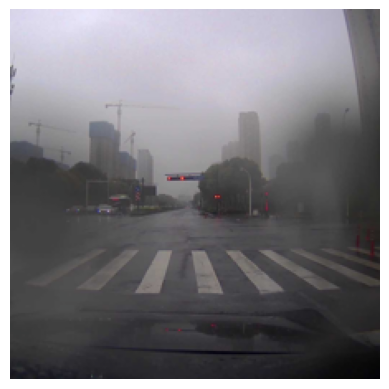} &
        \includegraphics[width=0.76 \linewidth]{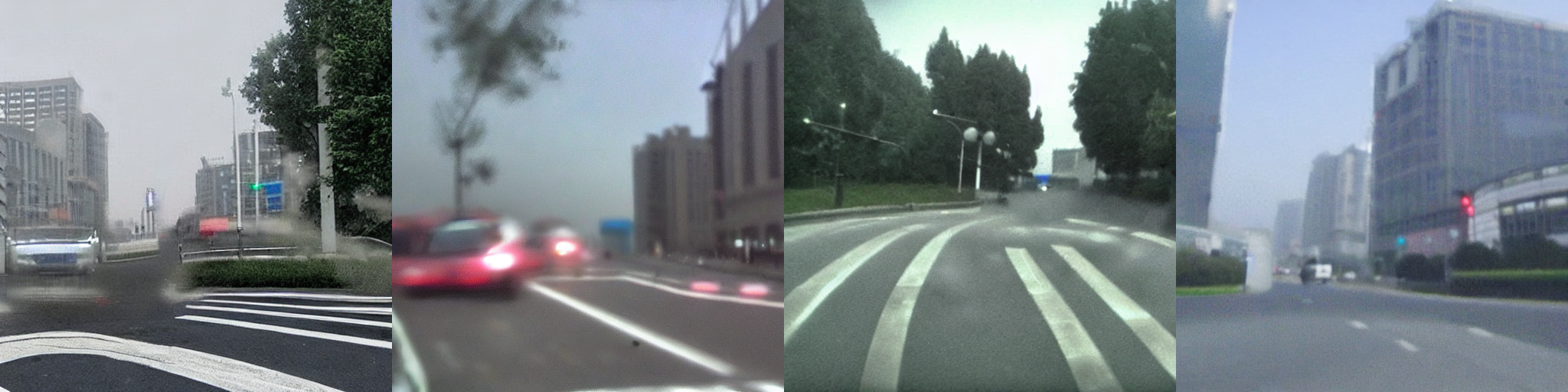} \\
        \multicolumn{2}{c}{Generated caption: A street with a row of parked motorcycles and a yellow fire hydrant.} \\
        \includegraphics[width=0.19 \linewidth]{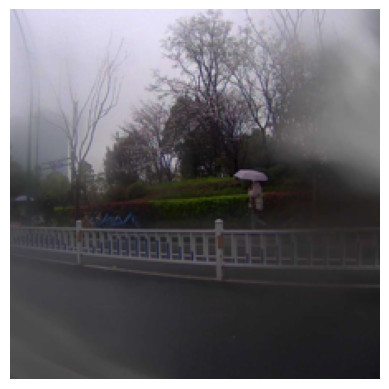} &
        \includegraphics[width=0.76 \linewidth]{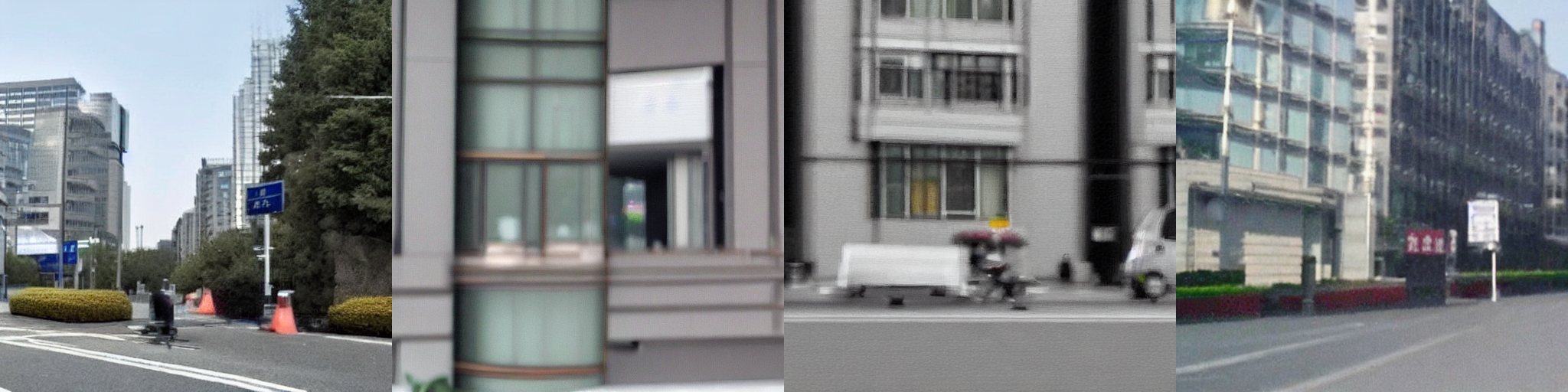} \\
        \multicolumn{2}{c}{Generated caption: A bus is driving down the street with a man standing on the sidewalk.} \\
        \includegraphics[width=0.19 \linewidth]{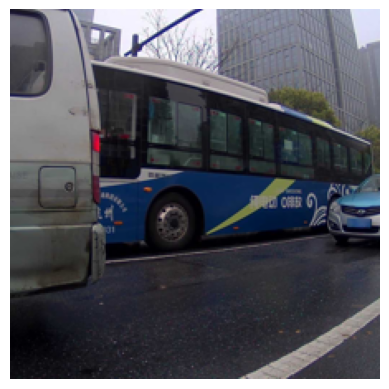} &
        \includegraphics[width=0.76 \linewidth]{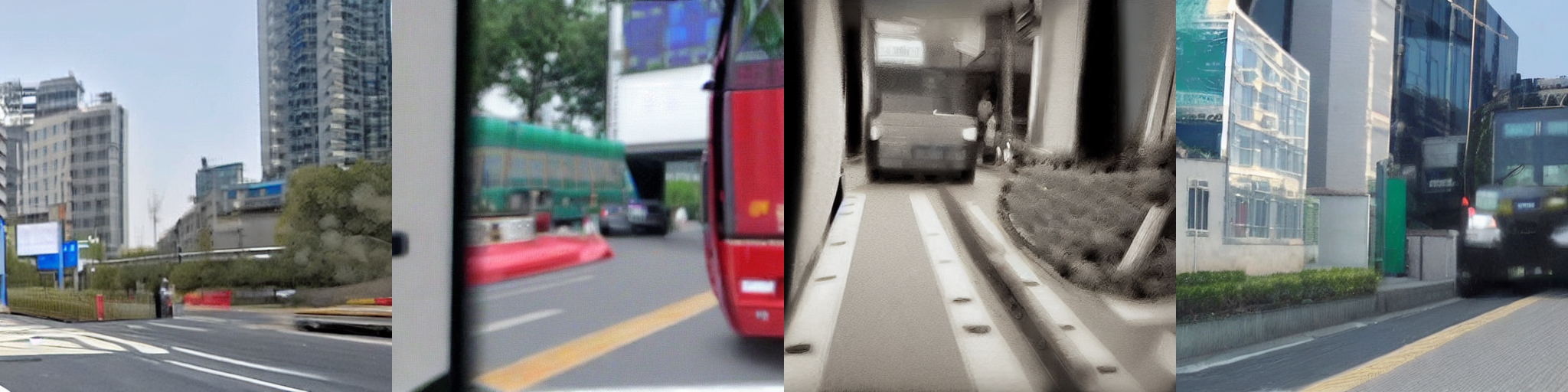} \\
        \multicolumn{2}{c}{Generated caption: A street with a lot of cars and a building.} \\
        \includegraphics[width=0.19 \linewidth]{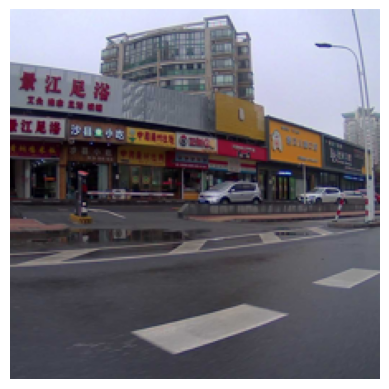} &
        \includegraphics[width=0.76 \linewidth]{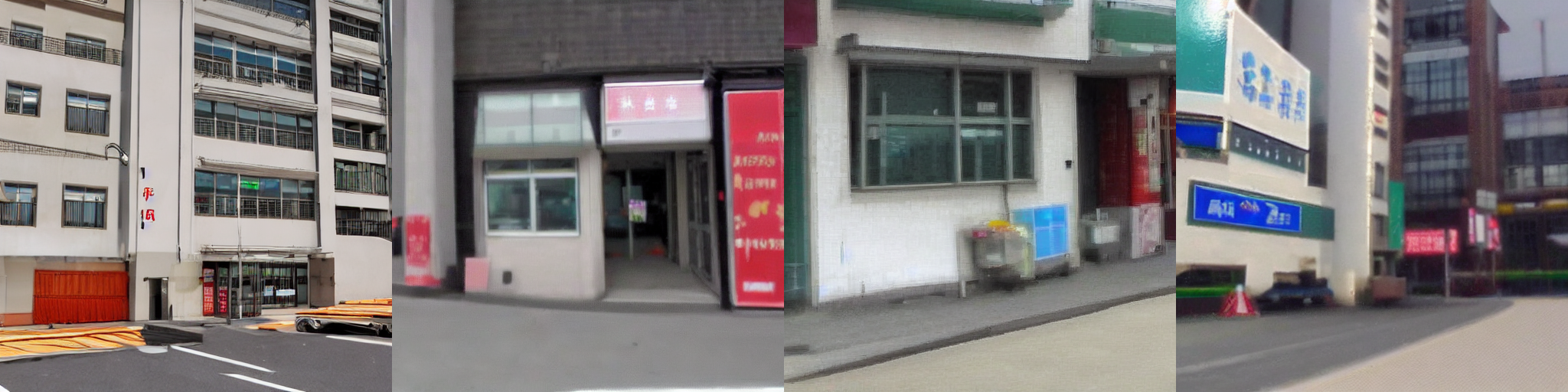} \\
        \multicolumn{2}{c}{Generated caption: A view of a highway with a red stoplight.} \\
        \includegraphics[width=0.19 \linewidth]{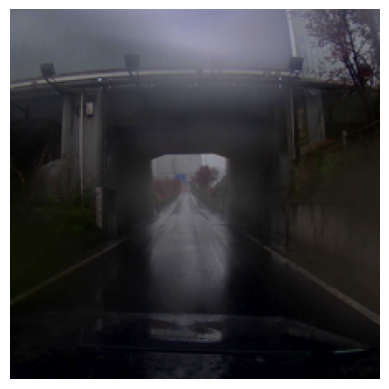} &
        \includegraphics[width=0.76 \linewidth]{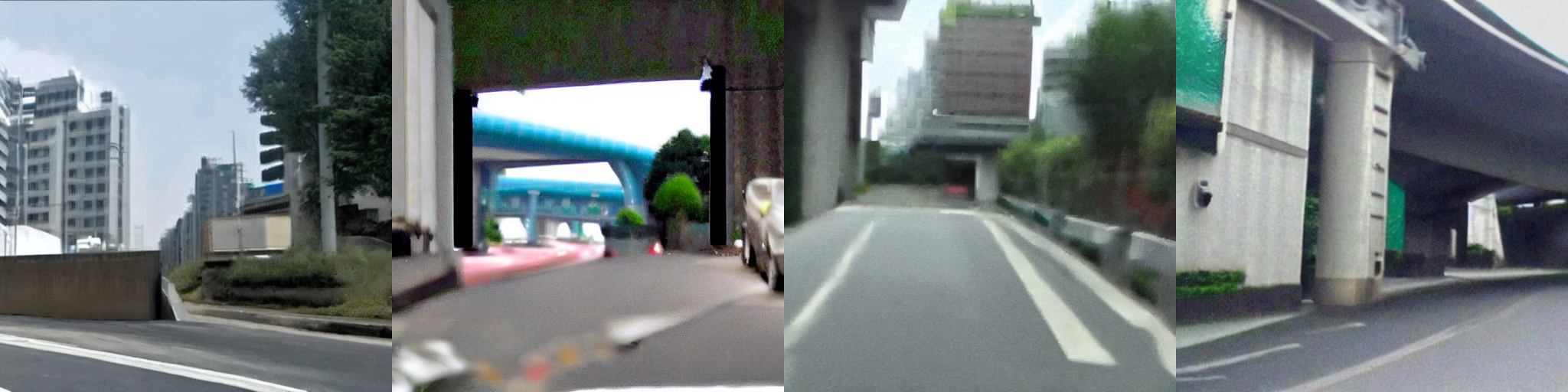} \\
    \end{tabular}
    \centering
    \caption{Example of generative application of LidarCLIP. A point cloud is embedded into the clip space (left, image only for reference) and used to generate text (top) and images (right). All four images are only generated with guidance from the lidar embedding, the caption was not used for guidance.}
    \label{fig:generative_suppl}
\setlength{\tabcolsep}{6pt}
\end{figure*}

We provide additional examples of generative applications of LidarCLIP in \cref{fig:generative_suppl}. These are randomly picked scenes with no tuning of the generative process. The latter is especially important for images, where small changes in guidance scale\footnote{The guidance scale is a parameter controlling how much the image generation should be guided by CLIP, which may stand in contrast to photorealism.} and number of diffusion steps have a massive impact on the quality of the generated images. Furthermore, to isolate the impact of our lidar embedding, we use the same parameters and random seeds for the different scenes. This leads to similar large-scale structures for images with the same seed, which is especially apparent in the rightmost column.

In most of these cases, the generated scene captures at least some key aspect of the embedded scene. In the first row, all generated scenes show an empty road with many road paintings. There is also a tendency to generate red lights. Interestingly, several images show localized blurry artifacts, similar to the raindrops in the source image. The second row shows very little similarity with the embedded scene, only picking up on minor details like umbrellas and dividers. In the third row, the focus is clearly on the bus, which is present in the caption and three out of four generated images. In the fourth row, the storefronts are the main subject, but the generated images do not contain any cars, unlike the caption. In the final scene, we see that the model picks up the highway arch in three out of the four generated images, but the caption hallucinates a red stoplight, which is not present in the source.

\end{document}